\documentclass{clv2025}
\usepackage{placeins}
\usepackage{float}
\usepackage{xurl}
\hypersetup{hidelinks}
\title{Divergent large language model predictions from convergent
representations in ambiguous word pairs}

\author{K. Jack Scott$^{1}$\thanks{Corresponding author:
\texttt{jack.scott@otago.ac.nz}}, Narun Pat$^{1}$, Veronica Liesaputra$^{2}$}

\affilblock{%
  \affil{Department of Psychology, University of Otago, Dunedin, New Zealand}%
  \affil{School of Computing, University of Otago, Dunedin, New Zealand}%
}

\pageonefooter{}

\begin{document}
\pagestyle{pageonly}
\maketitle
\thispagestyle{pageonly}

\begin{abstract}
In this work we investigate how decoder-only transformers resolve lexical
ambiguity through layer-by-layer analysis of three models spanning three
parameter sizes (GPT-2-Small-117M, Llama-3.2-3B, Qwen2.5-32B). For both
homonyms and polysemes, we find that representations become maximally distinct
in middle layers, then partially reconverge in late layers, while the KL divergence
between their next-token predictions reaches its maximum in the final layers.
The activation patching experiment provides causal evidence that
late-layer representational differences directly determine outputs despite
apparent increased similarity in embedding space. Our single-layer ablation
experiment indicates that models achieve equivalent disambiguation despite
qualitatively different layer-wise vulnerabilities. These findings offer a
mechanism for recent observations where models' internal embedding similarities
show low correlation with their behavioural outputs despite strong performance.
The semantic distinctions therefore remain present but become increasingly 
invisible to similarity measures over the embeddings, with implications for 
embedding-based methods such as semantic search, retrieval, and clustering that 
rely on late-layer cosine similarity.
\end{abstract}

\section{Introduction}

In most deployed natural language processing (NLP) systems, ``semantically
similar'' has an operational definition: two texts count as similar when their
extracted embeddings have high cosine similarity. Retrieval-augmented
generation, semantic search, and clustering often rest on the assumption of the
equivalence of geometric proximity and meaning \citep{patil2023}. Lexically
ambiguous words offer a direct test of it, since they require a model to
represent one surface form as several distinct meanings, and how well
late-layer embedding geometry captures those distinctions is not fully
established. There are reasons to doubt late-layer embedding geometry is a clean
capture. Contextualized representations can be highly anisotropic, occupying a
narrow cone in which even unrelated tokens can have high cosine similarity
\citep{ethayarajh2019,razzhigaev2024}. This anisotropy may be driven by a small
number of `rogue dimensions' that dominate the cosine computation and can
obscure genuine representational structure \citep{timkey2021}. As a result,
cosine similarity over learned embeddings can yield results that are opaque or
even arbitrary depending on how the underlying model was regularized
\citep{steck2024}. If methods in the field often rely on geometric distance to
locate semantic meaning, but that metric may be unreliable, then it is not
immediately obvious how or where in the network the information needed for
disambiguation is actually carried.

Homonyms and polysemes provide a clean natural experiment for that question.
Strict homonyms (identical words with unrelated meanings, such as ``bank'' as a
financial institution versus a riverbank) and concept-drift polysemes
(identical words with related but diverged meanings, such as ``mouse'' as a
rodent versus a computer device) require a model to construct distinct
representations from identical input tokens, isolating the use of context for
disambiguation while holding the surface form constant. They therefore offer a
controlled setting in which to ask whether late-layer embedding geometry
captures the semantic distinctions a model demonstrably acts on.

A recent finding by \citet{capone2024} brings this question into focus. The
researchers reported that instruction-tuned decoder models produced similarity
ratings for polysemous words that moderately tracked human judgments, yet those
same ratings correlated only weakly with the models' own internal embedding
similarities \citep{capone2024}. In other words, the models behaved as though
they distinguished senses that their embedding geometry did not clearly
separate. This points to a gap between how contextual information is represented
geometrically and how it is expressed behaviourally, but the mechanism behind
that gap has not been characterized. Prior work has examined end-task
performance \citep{proietti2024,yae2025}, and more recently, the layer-wise
trajectory of representational separation for ambiguous words
\citep{ma2025,riviere2024}, but without relating that trajectory to the model's
predictions or testing whether the late-layer representations causally drive
disambiguation.

We address this with an experimental design that follows disambiguation through
the network rather than reading it off a single layer. First, using the logit
lens \citep{nostalgebraist2020}, we measure, at every layer of three
decoder-only models spanning three orders of magnitude (GPT-2-Small [117M],
Llama-3.2-3B, Qwen2.5-32B), both how the models represent ambiguous words
geometrically (cosine distance between hidden states and between projected
logits) and how they predict the next token (KL divergence between output
distributions). This separates representational encoding from predictive
encoding. Second, we use activation patching \citep{heimersheim2024} to provide
causal evidence for which layers' representations actually determine the output.
Third, we use single-layer ablation to ask whether different models depend on
the same layers to maintain disambiguation. By spanning three model scales, two
ambiguity types, and both semantically unrelated and semantically related
meanings, we test whether any observed pattern is general or model-specific.

Three questions organize the study. First, does the geometry of a model's
late-layer representations, read through the linear similarity measures
embedding methods often rely on, expose the distinction between a word's senses,
or does that distinction become apparent only through the nonlinear map to the
output? Second, regardless of geometric structure, do these late-layer
representations still causally drive the model's ability to disambiguate? Third,
do models from different developers achieve disambiguation through the same
layer-wise dependencies?

We find a consistent answer across all three models and both ambiguity types:
representations of the same word in different contexts become maximally distinct
in middle layers and then partially reconverge in late layers, exactly where
prediction distributions diverge most strongly. Activation patching shows that
these late-layer representations, despite their reduced separation, causally
determine the output. The semantic distinctions are therefore preserved, but
increasingly through their influence on predictions rather than through
geometric separation. This has implications for embedding-based NLP: geometric
similarity over late-layer embeddings can understate the disambiguation a model
actually performs, and methods that draw on the model's predictions, not its
embedding geometry alone, may recover distinctions that cosine similarity
misses.

\section{Related Work}

\subsection{Lexical ambiguity in language models}

While modern language models perform highly on Word Sense Disambiguation (WSD)
tasks, gaps remain in our understanding of how the internal computational
pathways enable this performance. \citet{proietti2024} found that pretrained
encoder-decoder models distinguish homonyms reliably, with representations of
similar senses clustering together, and \citet{yae2025} reported that WSD
performance scales with parameter count. These works highlight that models can
disambiguate but characterize either final-layer representations or end-task
accuracy rather than the layer-wise computation that produces them.

Closest to our study, \citet{ma2025} computed a layer-wise angular
disambiguation score for homonyms across BERT, GPT-2, Llama 3, and Qwen 2.5 in
English and Chinese, and reported that the best layer for separating senses
varies by family, with non-monotonic trajectories that peak at different depths
(the Llama and Qwen families rise and then retrace in the late layers).
\citet{riviere2024} likewise found that the distance between an ambiguous word's
representations follows layer-wise trajectories that depend on architecture and
scale. These studies establish, in representation space, that homonym
separability is non-monotonic and model-dependent, using anisotropy-corrected
angular measures over pooled word representations. Neither examines how that
representational trajectory relates to the model's predictions, nor tests
whether late-layer representations causally drive disambiguation. Our study adds
those two elements, alongside a wider stimulus set that includes concept-drift
polysemes. We measure prediction space together with representation space and
find that representational reconvergence coincides with maximal predictive
divergence, and we use activation patching to show that the late-layer
representations remain causal for prediction despite their reduced separation
between homonym/polyseme pairs.

\subsection{Layer-wise geometry of representations}

A separate line of work characterises how representational geometry changes with
depth. \citet{ethayarajh2019} showed that a word's representations grow more
context-specific in upper layers, with self-similarity declining across depth.
\citet{razzhigaev2024} found that anisotropy in decoder models follows a
bell-shaped profile peaking in the middle layers, unlike the flatter encoder
profile, so the compression of the representation space is itself strongly
layer-dependent. \citet{queipo2025} trace a corresponding middle-layer
compression valley to massive activations in the residual stream, linking it to
attention sinks and to high anisotropy. \citet{skean2025} report that
intermediate layers often yield stronger features for downstream embedding tasks
than the final layer, challenging the typical approach of extracting from the
penultimate or final layer. These studies establish that geometric structure
varies substantially across depth, but they treat geometry as the object of
study; none connects the layer-wise geometric trajectory to the causal
computation that resolves a specific ambiguity, which is the link we draw.

\subsection{Layer-wise interpretability methods}

The logit lens \citep{nostalgebraist2020} projects intermediate representations
to vocabulary space to reveal how predictions evolve across depth.
\citet{belrose2023} noted that the raw logit lens can be brittle for some models
and introduced the tuned lens, a learned affine refinement; we use the standard
logit lens and discuss this choice in our limitations. Activation patching
\citep{heimersheim2024} enables causal interventions on internal activations,
and recent circuit-level methods such as cross-layer transcoders and attribution
graphs \citep{ameisen2025} trace computation through networks. These tools have
traced computation and representation, for example following multilingual
factual recall layer by layer and finding that one stage transfers across
languages while the other is language-bound \citep{fierro2025}. They have not,
however, been applied to the question we take up: whether late-layer
representational geometry still exposes a word's sense distinctions, or whether
those distinctions surface only through the nonlinear map to the output.

\section{Methods}

\subsection{Model selection and implementation}

We investigated three decoder-only transformer models spanning three orders of
magnitude in scale: GPT-2-Small (OpenAI; 12 layers, 117M parameters;
\citealp{radford2019}), Llama-3.2-3B (Meta; 28 layers, 3B parameters;
\citealp{grattafiori2024}), and Qwen2.5-32B (Alibaba Cloud; 64 layers, 32B
parameters; \citealp{yang2025}). These models feature different training
regimes, architectural choices (e.g.\ LayerNorm vs RMSNorm, different activation
functions) and tokenisation methods (GPT-2: Byte Pair Encoding (BPE),
\citealp{gage1994}; Llama: SentencePiece, \citealp{kudo2018}; and Qwen: tiktoken
Byte-level BPE, \citealp{jain2022}), and most clearly parameter scale, enabling
us to assess whether observed patterns reflect universal computational
principles or model-specific factors. We accessed these models through the
TransformerLens library (\citealp{nanda2022}; v2.1.6) running on PyTorch
(2.6.0), which provides a unified API for extracting activations and performing
interventions across different model architectures. All models were loaded
straight from TransformerLens, with no fine tuning or training taking place. No
model hyperparameters were altered from default during the course of this work.

We examined two types of lexical ambiguity: homonyms (words with unrelated
meanings: ``bank'' as financial institution vs riverbank) and concept-drift
polysemes (words with related but evolved meanings: ``mouse'' as rodent vs
computer device). Candidate words were selected based on: (1) multiple distinct
meanings in contemporary English, (2) common usage, (3) potential for
single-token representation in target models, (4) identical spelling across
contexts, and (5) capacity for natural sentence construction.

Homonyms were sourced from Grammar.com, Wikipedia, Merriam-Webster,
Dictionary.com, YourDictionary, StackExchange, and KnowledgeLost.org using
search terms ``strict homonyms list English identical spelling'' and ``complete
list homonyms same spelling.'' Concept-drift polysemes were sourced from TCK
Publishing, Medium, Wikipedia, CBS News, SF Gate, Reader's Digest, and
HowStuffWorks using search terms ``words changed meaning over time technology
internet digital'' and ``semantic shift examples diachronic polysemy words
evolved meaning.''

From these sources, we compiled a candidate set of 190 homonym pairs and 97
concept-drift polyseme pairs. Sentence contexts for these words were initially
generated using Claude Sonnet 4.5 (Anthropic) and subsequently edited by the
experimenters to satisfy experiment-specific constraints. This process yielded
two distinct stimulus sets optimized for different experimental requirements.

For logit lens experiments, we created contextually rich sentences emphasizing
semantic clarity without positional constraints. Example homonym pairs included
``The bank of the river was muddy'' vs ``I need to go to the bank to deposit
money,'' and ``The grizzly bear caught salmon'' vs ``She couldn't bear the
weight of her grief.'' Example concept-drift pairs included ``The cat chased the
mouse across the kitchen floor'' vs ``I need to replace the batteries in my
wireless mouse,'' and ``The dark cloud brings heavy rain'' vs ``The data cloud
stores my photos.'' After validation (see below), this yielded 156--167 homonym
pairs and 86--94 concept-drift pairs per model, varying due to model-specific
tokenisation and semantic boundary differences.

For activation patching experiments, we required identical tokenization and
token positions across context pairs, necessitating carefully engineered
parallel sentence structures such as ``I went to the bank for money'' vs ``I
went to the bank for fishing.'' These stringent positional constraints
substantially reduced viable pairs after validation, yielding 66--86 homonym
pairs and 25--36 concept-drift pairs per model (Table~\ref{tab:counts}; see
Supplementary Materials for complete stimulus lists).

\begin{table}
\caption{Validated stimulus counts by model and experiment}
\label{tab:counts}
\centering
\small
\begin{tabular}{lcccc}
\toprule
Model & \shortstack{Logit Lens\\Homonyms} & \shortstack{Logit Lens\\Concepts}
& \shortstack{Swap\\Homonyms} & \shortstack{Swap\\Concepts} \\
\colrule
GPT-2 Small  & 167 & 93 & 66 & 25 \\
Llama-3.2-3B & 166 & 94 & 74 & 32 \\
Qwen2.5-32B  & 156 & 86 & 86 & 36 \\
\botrule
\end{tabular}
\end{table}

All stimuli underwent validation to ensure meaningful contextual
differentiation. For each homonym or concept pair, we verified: (1) correct
tokenization, (2) consistent token position across contexts (activation patching
only), and (3) distinct next-token predictions between contexts. Context
distinctness was quantified via KL divergence between next-token probability
distributions, with thresholds of $D_{\mathrm{KL}} > 0.5$ for logit lens stimuli
and $D_{\mathrm{KL}} > 1.0$ for activation patching stimuli. Additionally,
activation patching stimuli required top-$k$ token overlap $< 0.4$ to ensure
strong semantic differentiation. Validation was performed separately for each
model to account for architecture-specific tokenization and semantic
boundaries.

\subsection{Experiment 1: Logit lens}

The logit lens \citep{nostalgebraist2020} projects intermediate layer
representations directly to vocabulary space, revealing how predictions evolve
across network depth. For each validated homonym/concept pair, we extracted the
hidden state $h_{l,t}$ at the target word position from layer $l$'s residual
stream, then projected it through the model's unembedding matrix $W_U$ to obtain
layer-specific logits. These logits were converted to probability distributions
via softmax.

For each layer and context pair, we computed three metrics:

\noindent\textbf{Activation Distance:} Cosine distance between raw hidden state
representations, measuring how differently the model represents the same word in
different contexts within activation space.

\noindent\textbf{Logit Distance:} Cosine distance between projected logit
vectors (pre-softmax), measuring representational differences after
transformation to vocabulary space.

\noindent\textbf{KL Divergence:} Kullback-Leibler divergence between next-token
probability distributions, quantifying how differently the model predicts
subsequent tokens. We implemented numerically stable KL divergence calculation
with epsilon smoothing to handle zero-probability cases.

Formally, the key metrics were computed as:

\noindent Cosine Distance:
\[
d_{\cos}(\mathbf{h}_A,\mathbf{h}_B) = 1 - \frac{\mathbf{h}_A \cdot \mathbf{h}_B}
{\|\mathbf{h}_A\|\,\|\mathbf{h}_B\|}
\]
where $\mathbf{h}_A$ and $\mathbf{h}_B$ denote hidden state vectors for the
target word in the two contexts. Values range from 0 (identical direction) to 2
(opposite direction), with higher values indicating greater representational
difference.

\noindent KL Divergence:
\[
D_{\mathrm{KL}}(P\,\|\,Q) = \sum_i P_i \log \frac{P_i}{Q_i}
\]
where $P$ and $Q$ represent probability distributions over the vocabulary for
the two contexts, and epsilon smoothing ($\epsilon = 1\times 10^{-10}$) was
applied to prevent division by zero. Higher values indicate greater
distributional divergence between predictions.

\subsection{Experiment 2: Activation patching}

Activation patching tests whether layer-specific representations causally
influence model outputs by directly manipulating internal activations
\citep{heimersheim2024}. For each validated pair, we performed bidirectional
interventions at each layer: extracting the activation vector from Context A and
inserting it into Context B's forward pass at the target position (and vice
versa), then allowing the model to continue processing through all subsequent
layers to generate final predictions.

We quantified intervention effects using:

\noindent\textbf{Token Overlap:} Proportion of top-$k$ predicted tokens
($k=10$) matching the source context, measuring whether swapped predictions
shift toward the grafted representation's context.

\noindent\textbf{Swap Effectiveness:} Ratio comparing KL divergence between
swapped and original distributions to KL divergence between the two baseline
contexts, capturing how completely predictions shift toward the source:
\[
\mathrm{SwapEffectiveness}_{A\rightarrow B} =
\frac{D_{\mathrm{KL}}(P_{A,\mathrm{swap}}\,\|\,P_{A,\mathrm{original}})}
{D_{\mathrm{KL}}(P_{A,\mathrm{swap}}\,\|\,P_{B,\mathrm{original}})}
\]
Higher values indicate more complete adoption of the source context's
predictions. Values near zero indicate the swap had minimal effect on
predictions, while large values ($\gg 1$) indicate strong causal influence.
Non-finite values represent complete adoption where swapped
predictions become identical to the source context.

Importantly, swap effectiveness measures the influence of layer-specific swaps
on final model outputs after processing through all subsequent layers, capturing
causal persistence across the network rather than merely local effects. For
visualization purposes, swap effectiveness summary statistics were calculated in
$\log_{10}$ space to account for the metric's multiplicative nature and
log-normal distribution across orders of magnitude. Geometric means and 95\%
confidence intervals were computed by calculating mean $\pm$ 1.96 $\times$ SEM
in log space and back-transforming to the original scale. This approach is
standard for ratio data and ensures confidence intervals respect the metric's
positivity constraint.

\subsection{Experiment 3: Single-layer ablation analysis}

Our single-layer ablation experiment investigated layer-wise dependencies of our
three logit lens metrics by zeroing individual layer residual stream activations
and measuring aggregate degradation across all subsequent layers. We implemented
multiple foundation sizes, protecting 0--1, 0--2, or 0--3 layers to ensure that
ablation effects reflected computational processing rather than disrupting
information flow from early embedding.

For each ablated layer, we measured degradation in activation distance between
stimuli pairs, logit distance between stimuli pairs, and KL divergence between
next-token distributions. Degradation was quantified as percentage reduction in
the metric across all subsequent layers relative to baseline (no ablation).

\subsection{Sequence position control}

To test whether activation-space separation reflected semantic disambiguation rather 
than sequence position or syntactic-role changes alone, we compared residual-stream cosine 
distances for homonym pairs with distances between matching token strings appearing in 
reordered sentence pairs (e.g.,
``cat'' in ``The cat chased the mouse'' vs ``The mouse chased the cat''). 
Residual-stream activations were extracted at each of the 28 
Llama-3.2-3B layers, and cosine distance was calculated between corresponding token 
representations. The control analysis was restricted to activation distance; logit 
distance and KL divergence were not calculated for the sequence-order stimuli.

\subsection{Statistical analyses}

To account for hierarchical data structure (multiple observations per homonym
across layers), we employed statistical approaches appropriate to each model's
characteristics. For GPT-2, we used linear mixed-effects models with layers as
fixed effects and homonyms/concepts as random effects, fitted via restricted
maximum likelihood. Model specification: metric $\sim$ layer\_group $+$
$(1\,|\,\text{stimulus\_id})$, where layer group was treated as a categorical
fixed effect and stimulus\_id as a random intercept accounting for repeated
measures across layers for each homonym or concept pair. Significance of fixed
effects was assessed via one-way ANOVA tests.

For Llama-3.2-3B and Qwen2.5-32B, where KL divergence values became non-finite
in final layers (violating parametric assumptions), we employed non-parametric
Kruskal-Wallis tests across all metrics. To enable cross-layer comparisons
despite extreme value ranges, we used normalized ranks (ranks divided by sample
size) rather than raw KL divergence values. These non-finite values arose where a 
late-layer next-token distribution became near-deterministic, that is, at the extreme 
of the divergence being measured, and were dropped from each stimulus's ranking rather 
than capped, so the reported late-layer divergence ranks are conservative with respect 
to the late-layer increase. Where Kruskal-Wallis tests indicated
significant differences, selected pairwise comparisons between pre-defined layer
groups were conducted using Mann-Whitney U tests.

Layers were grouped as early/middle/late based on model depth at an approximate
and descriptive level from observation of our early logit lens findings: GPT-2
(0--3/4--8/9--11), Llama-3.2-3B (0--6/7--16/17--27), Qwen2.5-32B
(0--14/15--48/49--63). For single-layer ablation, we used Kruskal-Wallis tests
and Mann-Whitney U tests. Linear mixed-effects models were attempted but failed
to converge for Llama and Qwen models, likely due to the extreme range of KL
divergence values in late layers combined with the hierarchical structure
(foundation size $\times$ layer position); in these instances, we report
Kruskal-Wallis test results.

All analyses were conducted in Python using statsmodels 0.14.4 (mixed-effects
models), scipy 1.15.2 (Kruskal-Wallis and Mann-Whitney U tests). Significance
threshold was set at $\alpha = 0.05$.
\nopagebreak
\section{Results}
\nopagebreak
\subsection{Experiment 1: Logit lens analysis}

The logit lens method projects intermediate layer representations to vocabulary
space, revealing how models transform contextual information across network
depth. We computed three metrics at each layer: activation distance (cosine
distance between hidden states), logit distance (cosine distance between
projected logits), and KL divergence (divergence between next-token probability
distributions). For GPT-2, we used linear mixed-effects models; for Llama and
Qwen, extreme KL divergence values in final layers (non-finite for
Llama) required non-parametric rank-based analyses.

\begin{figure}[H]
\centering
\includegraphics[width=\textwidth]{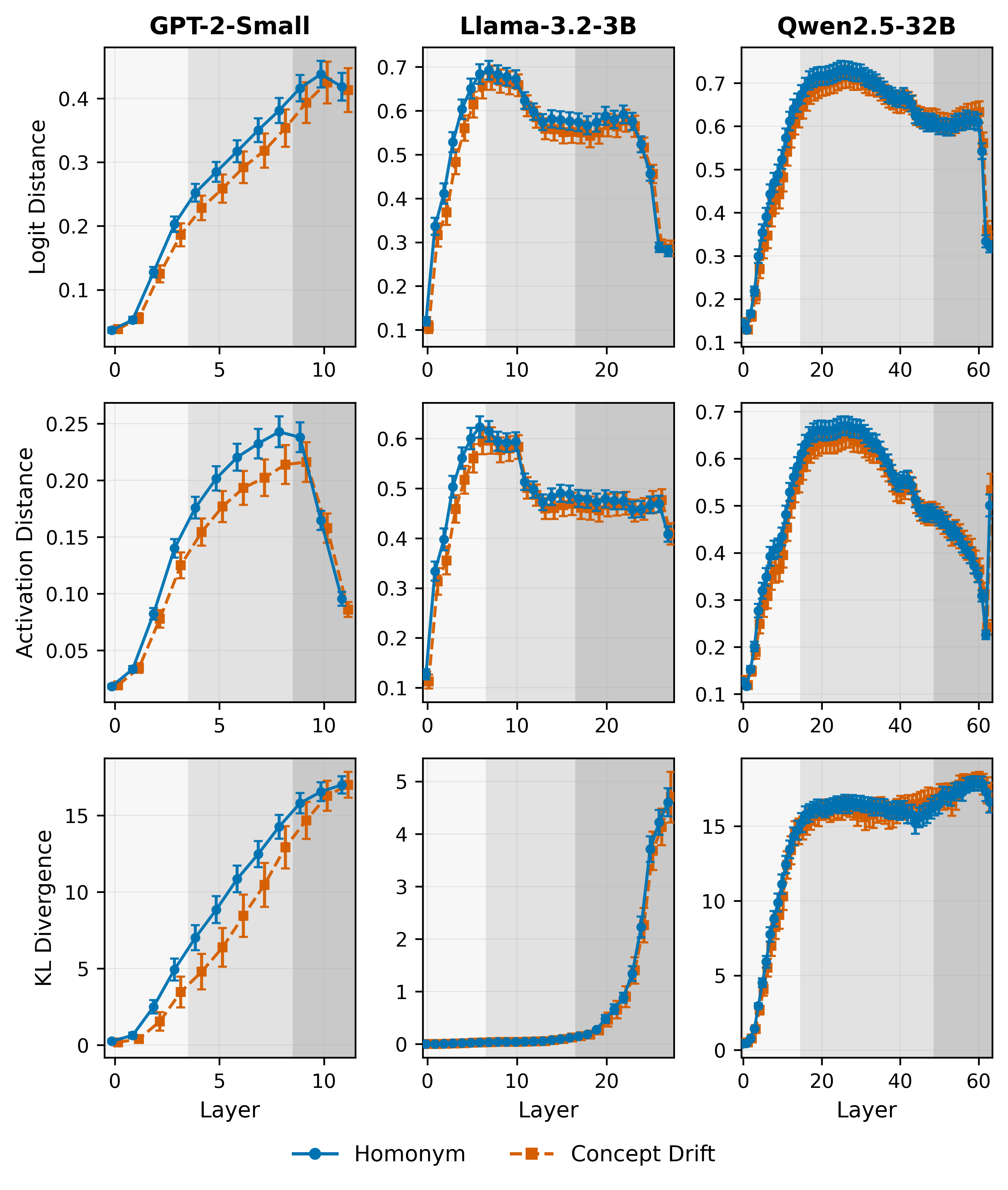}
\caption{Logit lens key distance metrics across layers. (A, B, C) Logit distance
between homonym and polyseme pairs across GPT-2-Small, Llama-3.2-3B and
Qwen2.5-32B models. (D, E, F) Activation distance across models. (G, H, I) KL
divergence (for prediction distributions) for homonyms and concept drift
polysemes.}
\label{fig:logitlens}
\end{figure}
\FloatBarrier
\subsubsection{GPT-2-Small}

For homonyms, activation distance showed a rise-peak-decline pattern
(Fig~\ref{fig:logitlens}D): early layers ($M = 0.069$) to middle layers
($M = 0.215$) to late layers ($M = 0.167$; $F(2, 2003) = 696.98$, $p < .001$).
Logit distance followed a similar trajectory in increasing distance across
layers, but did not reconverge late (Fig~\ref{fig:logitlens}A). KL divergence
similarly increased across layers: early $M = 2.08$, middle $M = 10.70$, late
$M = 16.45$ ($F(2, 1115) = 1318.39$, $p < .0001$).

For concept drift polysemes, activation distance showed a similar
rise-peak-decline pattern: early layers ($M = 0.064$) to middle layers
($M = 0.188$) to late layers ($M = 0.153$; $F(2, 1113) = 318.47$, $p < .001$).
Logit distance increased monotonically without reconvergence: early 0.101
$\rightarrow$ middle 0.290 $\rightarrow$ late 0.410 ($F(2, 1113) = 508.83$,
$p < .001$). KL divergence similarly increased: early 1.40 $\rightarrow$ middle
8.60 $\rightarrow$ late 16.00 ($F(2, 1113) = 550.12$, $p < .0001$).

The patterns are qualitatively identical to homonyms, though concept drift shows
slightly lower peak activation distance in middle layers and marginally lower
early-layer KL divergence.

\subsubsection{Llama-3.2-3B}

Llama demonstrated a similar pattern to GPT-2 in activation distance, though with
a more refined rise-peak-drop-plateau-decline (Fig~\ref{fig:logitlens}).
Activation distance increased from early layers (median $= 0.48$) to middle
layers (median $= 0.55$), then declined in late layers (median $= 0.48$;
$H = 329.07$, $p < 0.001$). Logit distance followed a similar pattern: early
0.51 $\rightarrow$ middle 0.65 $\rightarrow$ late 0.55 ($H = 575.91$,
$p < .0001$).

KL divergence fitted an exponential across layers, reflected in complete
separation between layer groups. Using normalized ranks due to infinite values
in final layers, early layers showed minimal divergence (median rank $= 0.14$),
middle layers intermediate divergence (0.46), and late layers maximum divergence
(0.82; $H = 3808.70$, $p < .0001$).

At a late layer (Layer 25), activation distance (0.48) was marginally lower than other
layers (0.51; $U = 317105$, $p = .001$), while KL divergence ranks were
significantly elevated (0.93 vs others 0.50; $U = 691363$, $p < .001$). Concept
drift polysemes showed near-identical patterns ($H = 152.46$ for activation
distance, $H = 2312.94$ for KL divergence; all $p < .0001$).

\subsubsection{Qwen2.5-32B}

Qwen's 64-layer architecture revealed the same essential pattern for activation
distance observed in GPT-2-Small and Llama-3.2-3B. Activation distance increased
from early layers (median $= 0.38$) to middle layers (0.63), then declined
substantially in late layers (0.42; $H = 3844.57$, $p < .001$), but
demonstrated a peculiar rise in activation distance in the final layer, which
may reflect transformations related to output projection. Effect sizes for layer
group transitions were large, with middle-to-late showing effect $= 0.87$. Logit
distance peaked later in the network (middle median $= 0.71$) with less dramatic
late-layer reconvergence (late median $= 0.61$; $H = 3232.93$, $p < .0001$).

KL divergence exhibited pronounced stratification across layer groups: early
0.14 $\rightarrow$ middle 0.53 $\rightarrow$ late 0.89 ($H = 5464.47$,
$p < .001$). Final layer analysis showed no significant activation distance
difference (layer 63: 0.50 vs others 0.53; $p = 0.28$) reflecting the steep
increase in distance observed in the final layer. While by contrast, KL
divergence ranks increased dramatically (0.92 vs others 0.52; $U = 1183155$,
$p < .0001$), demonstrating maximal prediction divergence despite
representational reconvergence.

These findings were replicated with concept drift polysemes with comparable
effect magnitudes (all $H > 1619$, $p < .001$).

Across all three models, representations of the same word in different contexts
became maximally distinct in middle layers, then partially reconverged in final
layers. Simultaneously, prediction distributions became increasingly divergent,
reaching maximum separation in late layers. This indicates that prediction
distribution divergence between pairs is not tightly coupled to geometric
distance of representations in late layers. By inference, the meaning carried in
the token is not distinguishable this way either in homonym and polyseme pairs.

\subsection{Experiment 2: Activation patching}

Activation patching tests whether layer-specific representations causally
influence model outputs. For each stimulus pair, we extracted the residual
stream activation from one context and inserted it into the other context's
forward pass at each layer, allowing the model to process through all subsequent
layers to generate final predictions. We measured token overlap (proportion of
top-10 predicted tokens matching the source context from which the activation
patch was taken) and swap effectiveness (KL divergence ratio between swapped and
baseline distributions).

\begin{figure}[H]
\centering
\includegraphics[width=\textwidth]{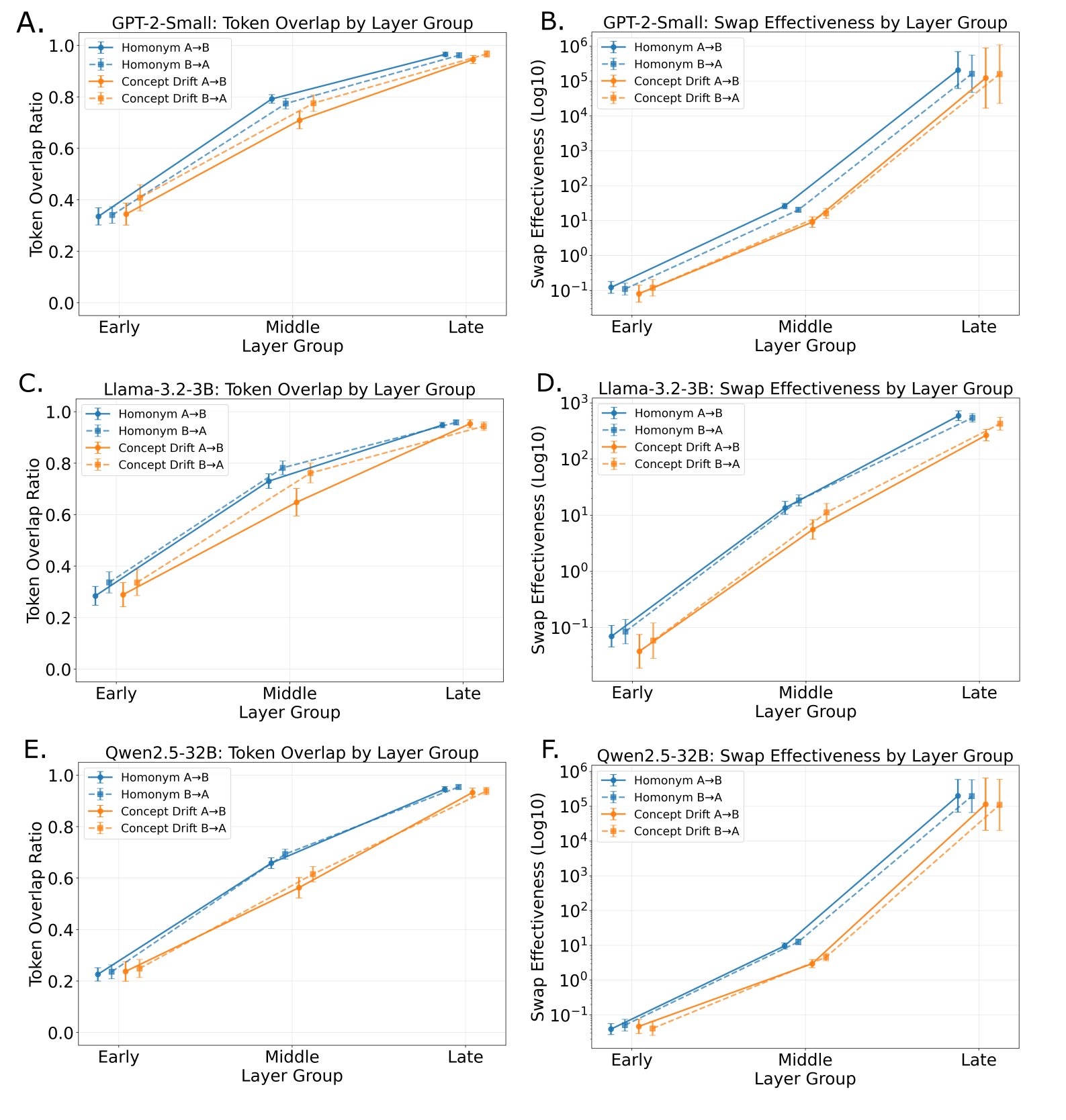}
\caption{Token overlap and swap effectiveness by layer group for GPT-2-Small
(A, B), Llama-3.2-3B (C, D), and Qwen2.5-32B (E, F), showing the progression from
minimal early effects to strong late-layer causal influence. For swap
effectiveness (B, D, F), points show geometric means with multiplicative 95\%
confidence intervals (calculated in $\log_{10}$ space; see Methods).}
\label{fig:patching}
\end{figure}
\FloatBarrier
\subsubsection{GPT-2-Small}

Token overlap increased systematically across layer groups for both homonyms and
concept drift. For homonyms, early layers produced minimal output shifts
($A\rightarrow B$: 0.34, $B\rightarrow A$: 0.34), middle layers showed
moderate-to-high overlap ($A\rightarrow B$: 0.79, $B\rightarrow A$: 0.77), and
late layers achieved near-complete prediction convergence ($A\rightarrow B$:
0.97, $B\rightarrow A$: 0.96; $F = 720.83$, $p < .001$). Concept drift
polysemes showed identical progression: early 0.34--0.41 $\rightarrow$ middle
0.71--0.78 $\rightarrow$ late 0.94--0.97 ($F = 196.50$, $p < .0001$;
Fig~\ref{fig:patching}A). Swap effectiveness showed similarly dramatic
increases: for homonyms, median values rose from 0.11--0.14 (early) to 22--25
(middle) to 820--1033 (late, $H > 590.65$, $p < .0001$), while for concept
drift, median values rose from 0.08--0.11 (early) to 12--22 (middle) to 653--1087
(late) (Fig~\ref{fig:patching}B, $H > 224.19$, $p < .0001$).

The bidirectional consistency of swap effects ($A\rightarrow B$ and
$B\rightarrow A$) indicates that final layer representations, despite appearing
similar in activation space, contain functionally distinct contextual
information that directly determines model outputs.

\subsubsection{Llama-3.2-3B}

Llama demonstrated strong causal effects with similar progression. For homonyms,
token overlap increased from early layers ($A\rightarrow B$: 0.28,
$B\rightarrow A$: 0.34) through middle layers ($A\rightarrow B$: 0.73,
$B\rightarrow A$: 0.78) to late layers ($A\rightarrow B$: 0.95, $B\rightarrow A$:
0.96; $H = 338.94$, $p < .0001$). Concept drift polysemes showed comparable
patterns: early 0.29--0.34 $\rightarrow$ middle 0.65--0.76 $\rightarrow$ late
0.94--0.95 ($H = 142.01$, $p < .0001$). Swap effectiveness mirrored this
progression for homonyms (early 0.05--0.11 $\rightarrow$ middle 14--24
$\rightarrow$ late 621--677; $H = 384.59$ and 390.18, $p < .001$) and concept
drift (early 0.04--0.05 $\rightarrow$ middle 6--10 $\rightarrow$ late 306--424;
$H = 171.32$ and 171.41, $p < .0001$).

Interestingly, while Llama's logit lens analysis showed extreme KL divergence in
late layers, activation patching revealed that mid-to-late layer
representations (not just final layers) possess substantial causal influence over
outputs. Late layer swaps produced 95\%+ token overlap despite moderate
representational distances, suggesting efficient encoding of contextual
distinctions.

\subsubsection{Qwen2.5-32B}

Qwen's 64-layer architecture revealed the most gradual but ultimately complete
causal progression. For homonyms, token overlap increased from early layers
($A\rightarrow B$: 0.23, $B\rightarrow A$: 0.24) through middle layers
($A\rightarrow B$: 0.66, $B\rightarrow A$: 0.69) to late layers
($A\rightarrow B$: 0.95, $B\rightarrow A$: 0.95; $H = 636.35$, $p < .001$).
Concept drift showed similar patterns with slightly lower middle-layer effects:
early 0.24 $\rightarrow$ middle 0.56--0.61 $\rightarrow$ late 0.93--0.94
($H = 237.77$, $p < .0001$). Swap effectiveness showed the most dramatic
late-layer effects: for homonyms, values increased to 965--1298 in late layers
($H = 720.54$ and 736.78, $p < .001$), while concept drift reached 468--496
($H = 307.74$ and 324.79, $p < .0001$).

Qwen's extended depth allowed observation of more gradual accumulation of causal
influence across layers compared to shallower models, yet the endpoint remained
identical: near-complete output determination by late-layer representations.

\subsection{Experiment 3: Single-layer ablation}

The single-layer ablation experiment revealed which layers were computationally
important for disambiguation and output, exposing key architectural differences
despite the shared representational pattern observed in experiment 1. We zeroed
out activations at a given layer, and measured degradation in subsequent layers'
ability to maintain semantic separation.

\subsubsection{GPT-2-Small}

GPT-2 showed near complete degradation for both activation and logit distance
degradation across the first half of layers, declining precipitously from layer
6 onwards. Activation distance degradation declined from $\sim$100 (early
layers) through middle layers to $\sim$10--11.4\% (late layers), while logit
distance followed a similar trajectory from 100\% to $\sim$8.4--8.8\% (activation
distance homonym: $H = 1009.98$, concept drift: $H = 471.18$, $p < .00001$;
logit distance homonym: $H = 1149.63$, concept drift: 540.8484, both
$p < .0001$). KL divergence exhibited a similar pattern with early and middle
layers both exhibiting 100\% degradation, declining to 32.3\% in homonyms, and
74.8\% for concept drift (homonym $H = 376.91$, concept drift 442.15, both
$p < .00001$). Of all foundation-size effects tested for GPT-2-Small, only 
KL-divergence degradation for homonyms showed a significant effect of foundation size ($H = 7.86$, $p = .020$). 
Post-hoc comparisons indicated slightly greater degradation with a foundation size 
of 4 than 2, although the effect was small.The differential between
homonyms and concept drift in degradation during late layers may be attributable
to the age and scope of the training set for GPT-2 \citep{radford2019}.

The overall pattern in GPT-2 suggests early layers are critical for establishing
prediction distributions, while late layers serve primarily to refine
representations rather than determine outputs.

\subsubsection{Llama-3.2-3B}

Llama demonstrated a clear difference between representational and
predictive vulnerabilities. Activation distance and logit distance degradation
remained consistently low throughout the network. Activation distance
degradation ranged from 7.9--8.3\% in early layers, peaking at 14.2--14.3\% in
middle layers, before declining to 4.2--4.1\% in late layers (homonym:
$H = 2026.21$, concept drift: $H = 1390.99$, both $p < .0001$). Logit distance
followed a similar pattern, ranging from 7.2--7.4\% in early layers to
14.5--14.2\% in middle layers and declining to 4.9--8.2\% in late layers
(homonym: $H = 1122.89$, concept drift: $H = 1201.07$, both $p < .0001$).

In sharp contrast, KL divergence degradation remained near-complete across all
layer groups, showing minimal decline even in late layers: early 100\%, middle
100\%, late 95\% for both homonyms and concept drift (homonym: $H = 6892.78$,
concept drift: $H = 3564.90$, both $p < .0001$).

These divergent patterns in degradation between metrics indicate that while 
coarse representational distinctions between contexts remain largely intact after 
ablating individual layers, precise prediction differentiation collapses. 
Unlike GPT-2 and Qwen where representations and predictions degrade together, 
Llama maintains robust representations while predictions remain highly 
vulnerable throughout the network.

Foundation-size effects were significant for all three homonym metrics: 
activation-distance degradation ($H = 19.65$, $p <.001$), logit-distance degradation ($H = 43.51$, $p <.001$), and KL-divergence degradation ($H = 81.55$, $p <.001$). For concept-drift stimuli, only KL-divergence degradation showed a significant foundation-size effect ($H = 21.03$, $p < .001$). In each of these analyses, larger protected foundations were associated with greater degradation, with KL-divergence degradation increasing monotonically across foundation sizes 2, 3, and 4.

\subsubsection{Qwen2.5-32B}

\begin{figure}[H]
\centering
\includegraphics[width=\textwidth]{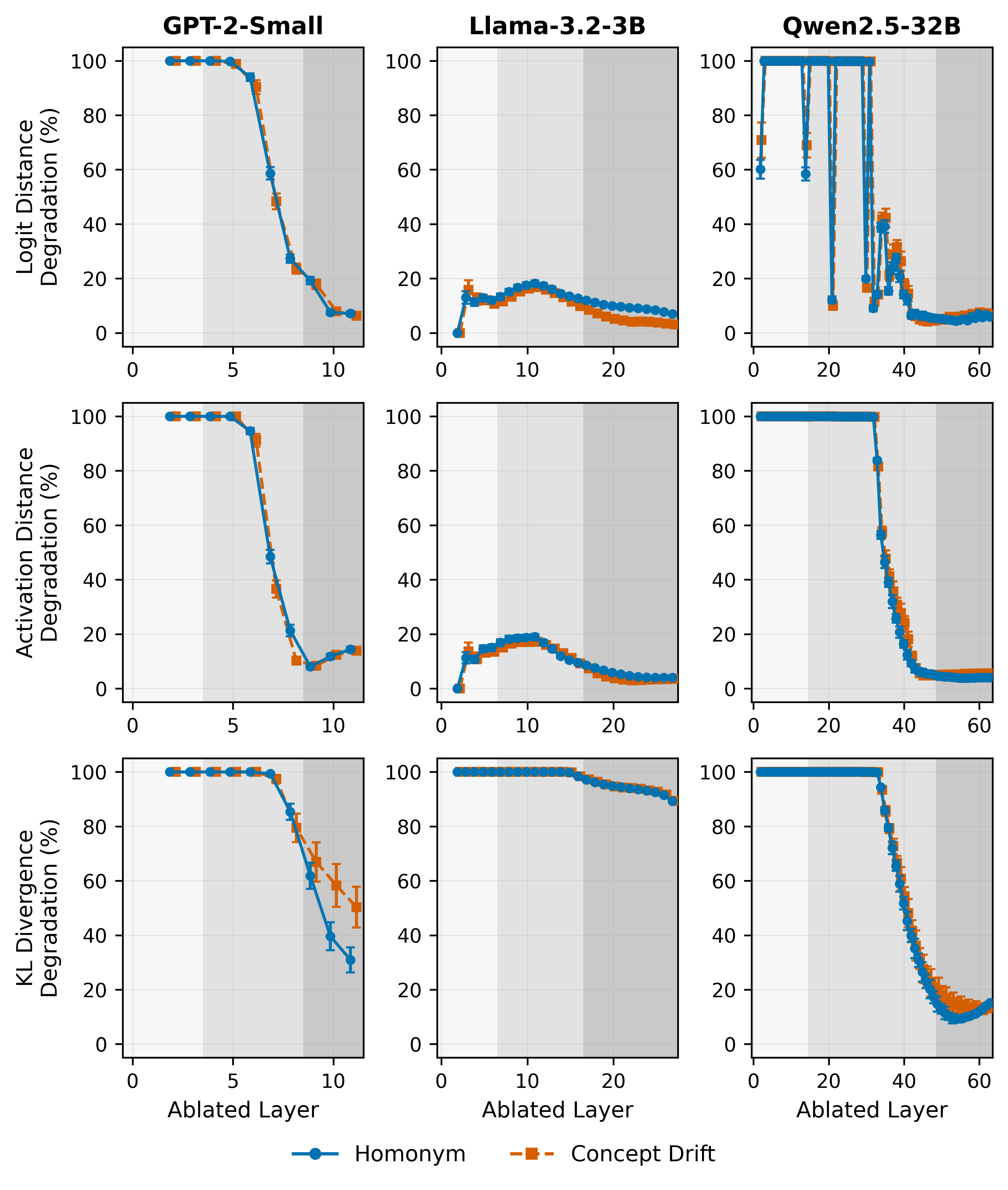}
\caption{Degradation percentage following single-layer ablation.}
\label{fig:ablation}
\end{figure}

Qwen's 64-layer architecture showed high degradation across all metrics until
layer 30, where degradation declined precipitously, in a manner not dissimilar
to GPT-2-Small. Activation distance degradation declined from 100\% (early
layers) and 99.9\% (middle layers) to $\sim$4.6--5.3\% (late layers; homonym
$H = 8074.62$, $p < .0001$; concept drift $H = 4388.89$, $p < .0001$). KL
divergence showed a distinctive two-phase pattern: degradation remained at 100\%
until approximately layer 30, then declined smoothly to approximately 10\%
(homonym $H = 8625.90$, $p < .0001$; concept drift $H = 4388.89$, $p < .0001$),
with a slight uptick in final layers for homonyms (Fig~\ref{fig:ablation}I).
Logit distance exhibited discontinuous drops at specific layer ranges
(Fig~\ref{fig:ablation}C), but generally declined from 100\% through early
layers, to 35--38\% in middle layers, to 4.5--6.3\% in late layers (homonym
$H = 7171.94$, $p < .0001$, concept drift $H = 3954.54$). The discontinuous and
pronounced drops in degradation for logit distance potentially reflect layers
robust to degradation, or alternatively numerical instabilities in ablated
residual streams.

Foundation size effects were significant for logit distance in both homonyms
($H = 20.82$, $p < .001$) and concept drift ($H = 14.16$, $p = .001$), but not for
activation-distance or KL-divergence degradation. Homonym logit-distance degradation 
peaked at foundation size 3,  whereas concept-drift logit-distance degradation was
lower at foundation size 4 than at sizes 2 and 3.

\FloatBarrier

Despite all models achieving similar representational reconvergence with
predictive divergence, there appears to be variation in how models accomplish
disambiguation, evidenced by qualitatively different degradation profiles across
metrics between models. GPT-2 showed a steep decline for all three metrics past
layer 6, indicating greater early-layer importance for predictions. Llama
exhibited decoupled vulnerabilities where representations remained robust, but
predictions collapsed. Qwen demonstrated the most distributed processing with
gradual degradation curves across its 64 layers. All models showed highly
significant layer effects (all $H$ or $F > 122$, $p < .0001$). Foundation-size 
effects were comparatively limited and model-specific. Llama showed the clearest 
monotonic increase in degradation with larger protected foundations, whereas GPT-2 
showed only a small homonym KL-divergence effect and Qwen showed non-monotonic 
logit-distance effects. For full coverage refer to supplementary materials Table S6. 

\subsection{Control analysis}

To test whether the observed activation-space separation could be explained by changes in 
sequence position or syntactic role alone, we compared Llama-3.2-3B activation distances 
for homonym pairs with those for matching tokens appearing in reordered sentence pairs. 
Homonym activation distances were substantially larger than sequence-order distances in early 
layers (median $= 0.483$ versus $0.113$, $U = 1{,}287{,}007.5$, $p < .001$, $r_{\mathrm{rb}} = .758$), 
middle layers (median $= 0.546$ versus $0.223$, $U = 2{,}839{,}429.0$, $p < .001$, $r_{\mathrm{rb}} = .901$), 
and late layers (median $= 0.482$ versus $0.190$, $U = 3{,}403{,}450.0$, $p < .001$, $r_{\mathrm{rb}} = .883$). 
Homonym distances were larger at every individual layer, indicating that the observed 
activation-space separation cannot be explained by sequence-order differences alone.

\begin{figure}[H]
\centering
\includegraphics[width=\textwidth]{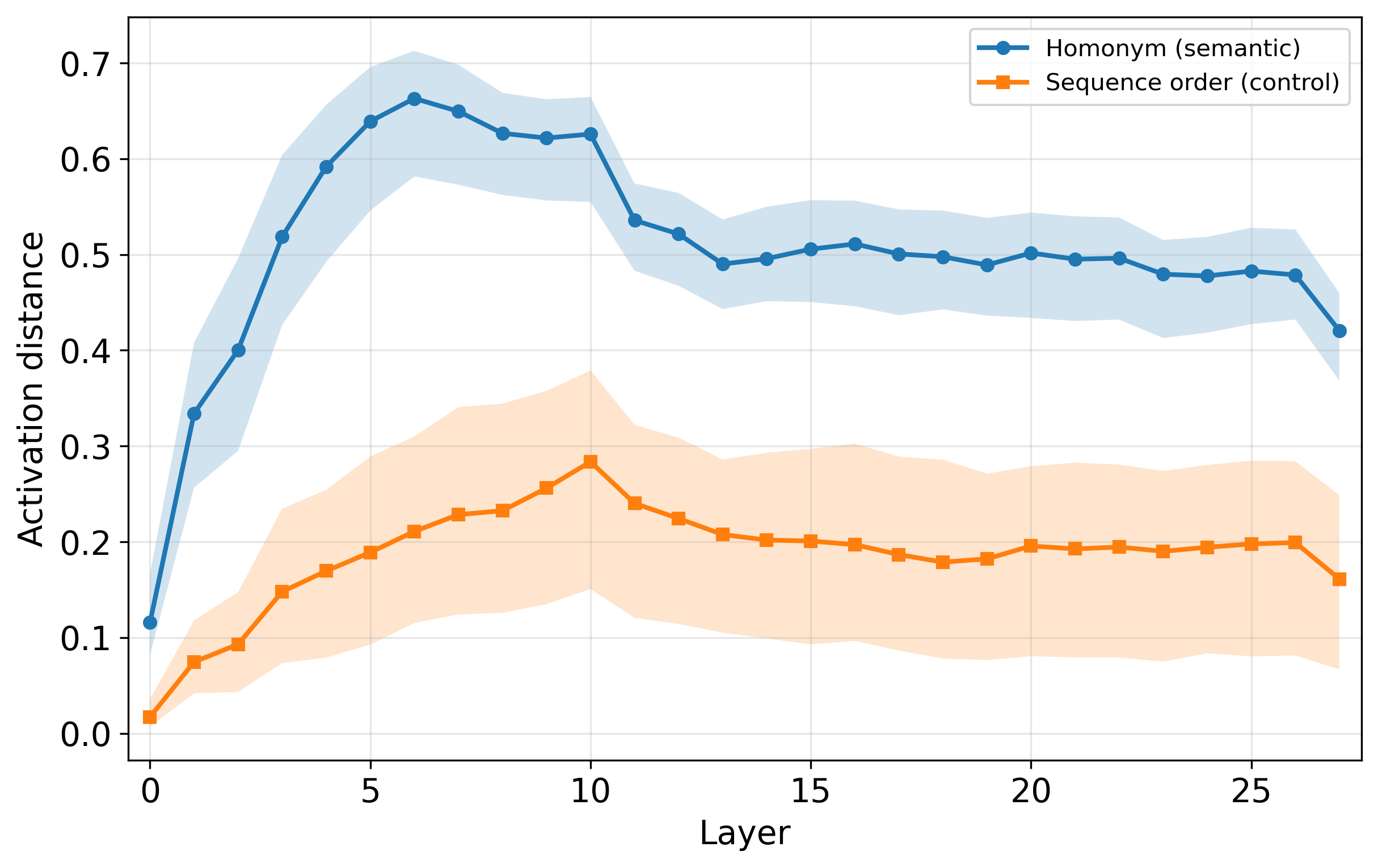}
\caption{Sequence position across layers compared with homonym pairs in
Llama-3.2-3B. Activation distance plot represents median values; shaded areas around 
lines represent Interquartile Range.}
\label{fig:control}
\end{figure}
\FloatBarrier

Across all three experiments, homonyms and concept drift polysemes showed
qualitatively identical patterns. Layer-wise trajectories, causal influence
profiles, and ablation vulnerabilities did not differ meaningfully between
stimulus types, suggesting the observed computational mechanisms are general to
lexical ambiguity resolution rather than specific to semantically unrelated
versus related meanings.

\section{Discussion}

We traced how three decoder-only transformers, spanning 117M to 32B parameters,
process lexically ambiguous words across network depth, with a focus on what
this implies for the late-layer embeddings used in NLP. Three findings held
consistently across models and across both homonyms and concept-drift
polysemes. First, cosine distance between the hidden-state representations of a
word in its two contexts rose through early and middle layers and then partially
reconverged in late layers in all three models; the same reconvergence appeared
in logit space for Llama and Qwen, though in GPT-2 logit-space distance rose
linearly. Second, predictive divergence between the two contexts (KL divergence)
increased monotonically, reaching its maximum in late layers. Third,
activation patching showed that late-layer representations causally determined
the output, with token overlap approaching 95\% when late-layer activations were
swapped between contexts. Control analyses confirmed that these effects
reflected semantic disambiguation rather than syntactic position.

The central result is that models reached maximum predictive separation
precisely where geometric separation between the representations was smallest.
The disambiguating information is not lost at that point; it is increasingly
carried in a form that angular distance between representations does not
sufficiently register, while remaining expressed in the output. Because each
ambiguous token is embedded in a disambiguating sentence, its representation has
already integrated the relevant context; the finding is therefore not that
context disambiguates, which is expected, but that the geometric separation
between the senses partially reconverges in late layers even as the distinction
remains causally verifiable. This may reframe what late-layer embeddings can and
cannot tell us. \citet{ethayarajh2019} established that contextualisation
increases with depth, with representations of a word becoming more
context-specific in upper layers. Our paired-context analysis qualifies that
trajectory, where for ambiguous words the geometric separation is non-monotonic,
peaking in middle layers and then partially reconverging late, even as the
functional distinction continues to sharpen. This trajectory also runs against
the layer-wise anisotropy profile of decoder models. \citet{razzhigaev2024}
report that anisotropy in decoders peaks in the middle layers, and
\citet{queipo2025} trace a corresponding middle-layer compression valley in the
residual stream to massive activations. They find that where the global
representation space is most compressed, the cosine distance between arbitrary
tokens is smallest. Yet, it is precisely there that the distance between an
ambiguous word's two senses is largest in the present work. The separation we
measure may thus be anti-correlated with the global geometry, which lends to the
argument that our finding reflects sense content rather than the anisotropy
profile, and that raw cosine distances, if anything, may understate the
mid-layer separation. The reconvergence does not seem to be a loss of
information. Activation patching shows the late-layer representations, despite
their reduced separation, still carry the disambiguating signal causally, and
the bidirectional symmetry of the swap effects, where A to B and B to A produced
comparable shifts, indicates this is genuine encoding rather than noise
reduction. The distinction is preserved, but it is increasingly expressed
through how representations map to predictions rather than through their
separation in embedding space.

This is the mechanism we propose to explain the decoupling phenomenon reported
by \citet{capone2024}. Their finding that behavioural similarity judgments
correlate weakly with internal embedding similarities is what one would expect if
a model's representational differences reconverge while its predictive
differences are amplified. More broadly, it offers a concrete reading of why
geometric similarity over contextual embeddings can be an unreliable proxy for
meaning \citep{ethayarajh2019,timkey2021,steck2024}. In the late layers from
which embeddings are typically extracted, the very distinctions a downstream task
cares about may have transformed in a manner not clearly captured by the metric.
For applications requiring fine-grained semantic discrimination, methods that
combine embedding similarity with information from the model's predictions may
capture more of a model's disambiguation capacity than geometry alone. On a
related note, because the geometric separation we observe is concentrated in
middle layers, extracting representations there rather than at the final layer,
which \citet{skean2025} find improves embedding quality across a range of tasks,
would better preserve the sense distinctions for similarity-based use. We frame
these as directions rather than results, since we do not evaluate a retrieval or
WSD system directly (see limitations).

The single-layer ablation experiment indicates that this shared outcome is
reached through different layer-wise dependencies across architectures. This
model-specificity is consistent with \citet{ma2025}, who found that no single
layer depth best separates homonym senses across models; our ablation adds that
these dependencies are causal and that they reveal differences between
representational and predictive degradation. Layer-intervention work has shown
that LLMs retain most of their next-token accuracy when middle layers are
deleted or swapped, motivating accounts of universal inference stages
\citep{lad2025}; for disambiguation we instead find model-specific layer
dependencies. GPT-2-Small and Qwen2.5-32B showed severe degradation of
representational and predictive metrics when early-to-middle layers were
ablated, declining sharply thereafter, whereas Llama-3.2-3B showed divergent
trajectories, where coarse representational separation (activation and logit
distance) was robust to single-layer ablation, while predictive differentiation
(KL divergence) remained near-completely degraded throughout the network. We
could not identify a clean architectural explanation. Llama and Qwen share
RMSNorm, grouped-query attention, rotary positional encoding, and SwiGLU
activations, yet group separately from each other behaviourally, while GPT-2 and
Qwen group together despite differing on all of these. What distinguishes Llama
here remains open. We report this as a robust descriptive difference between
models and note that we did not statistically test the between-model
differences, as these models had different layer depths, and scale of
degradation metrics, making like-for-like comparison ill defined. Qwen
additionally showed isolated layers where logit-distance degradation dropped
sharply, which may reflect genuine layer specialization at scale or numerical
instability under ablation; we cannot adjudicate between these here.

\subsection{Limitations and future directions}

Several limitations bound these claims. Our stimuli are single-token
residual-stream activations in base decoder models, whereas production embedding
and retrieval systems typically use pooled sentence or document vectors, often
from models fine-tuned specifically for embedding. The bridge to applied
cosine-similarity failure is therefore suggestive rather than direct, and our
claims are scoped to how base decoder LMs encode lexical disambiguation across
depth, with implications for, rather than a demonstration of, downstream
embedding behaviour. We also report raw cosine distances without subtracting an
anisotropy baseline, unlike \citet{ma2025} and \citet{ethayarajh2019}; our
conclusions do not turn on this, since activation patching is geometry-free and
the decoder anisotropy profile runs opposite to the separation trajectory we
observe, but an anisotropy-corrected replication would make our geometric
measurements directly comparable to that work. 

We used the standard logit lens, which can be brittle for some models
\citep{belrose2023}; We did not replicate with the tuned lens, since recent work 
suggests the tuned lens's training objective is correlational rather than causal 
and tends to bypass unverbalised intermediate computation in favor of the eventual 
output, making it less suited than the logit lens to exposing intermediate 
representational content \citep{gurnee2026}. Our activation-patching results, 
which do not depend on either lens, corroborate the central finding causally 
regardless of this choice.

Our metrics also leave open whether the sense distinction is linearly decodable 
from the representation at a given layer: cosine distance between a pair of contextual 
representations indexes angular separation rather than linear separability, 
and activation patching establishes causal sufficiency through the full 
nonlinear remainder of the network rather than a linear readout at the patched layer.
An established layer-wise probing literature has localized sense and semantic-class
information, but largely in encoder models and largely to establish its presence 
rather than the late-layer geometric reconvergence we report; a trained probe at
each layer of our decoder models would test decodability directly, 
and we leave it to future work. We examined only decoder-only architectures 
on a single disambiguation task; whether the pattern generalizes to encoder or 
encoder-decoder models, or to syntactic, referential, and pragmatic ambiguity, is an
open question. Finally, our cross-architecture comparison cannot separate 
scale from architecture; a single model family evaluated across a full parameter 
sweep would isolate the effect of scale.

Recent circuit-level methods \citep{ameisen2025} now make it feasible to ask
which features, at which layers, drive the reconvergence of representational
separation alongside the amplification of predictive distinctions. Identifying
those circuits is a natural next step and would convert the descriptive pattern
reported here into a mechanistic account.

\subsection{Conclusion}

In sum, we find a consistent computational pattern across model scales,
architectures, and ambiguity types: representations of ambiguous words converge
in late layers while prediction distributions diverge, with the disambiguating
information preserved causally in those late-layer representations despite their
reduced separation. This clarifies why internal embedding geometry can diverge
from behaviour and cautions against treating late-layer cosine similarity as a
complete readout of the semantic distinctions a model makes. For practitioners
who rely on late-layer embeddings in tasks such as semantic search, retrieval, or
clustering, these results suggest that similarity computed over final-layer
geometry can understate the sense distinctions a model encodes, and that
combining embedding similarity with information from the model's predictions, or
drawing on middle-layer representations, may be better positioned to recover
distinctions that late-layer cosine similarity may miss.

\section*{Data and Code Availability}

The complete stimulus sets are provided in the Supplementary Materials. Code
required to reproduce the experiments, statistical analyses, and figures
reported in this study is available at
\url{https://github.com/scoki211/Divergent_LLM_Predictions_Convergent_Reps_amb_words}.

\bibliographystyle{compling}
\bibliography{references}

\end{document}


\pagestyle{pageonly}
\thispagestyle{pageonly}

\section*{Supplementary Materials}
\subsection*{Full homonym and concept drift polyseme tables}
Tables S1 to S4 list the full candidate pools of homonym and concept-drift polyseme pairs from which the per-model stimulus sets in Scott et al. (2026) were drawn. Tables are organised by experiment (logit lens: Tables S1 to S2; activation patching: Tables S3 to S4) and by ambiguity type (homonyms: Tables S1, S3; concept-drift polysemes: Tables S2, S4). Each table includes a Models used column indicating which model or models, GPT-2, Llama, or Qwen, retained that pair after per-model validation. A blank entry indicates the pair did not pass validation for any model.
\clearpage
\noindent{\bfseries Table S1}\par
\noindent Logit lens homonym stimuli: candidate pool of homonym pairs and sentence contexts, prior to per-model validation\par\smallskip
{\fontsize{6.7}{8.0}\selectfont
\begin{longtable}{@{}>{\raggedright\arraybackslash}p{0.08\linewidth}>{\raggedright\arraybackslash}p{0.185\linewidth}>{\raggedright\arraybackslash}p{0.185\linewidth}>{\raggedright\arraybackslash}p{0.185\linewidth}>{\raggedright\arraybackslash}p{0.185\linewidth}>{\raggedright\arraybackslash}p{0.12\linewidth}@{}}
\toprule
\textbf{Word} & \textbf{Context 1} & \textbf{Context 2} & \textbf{Context 3} & \textbf{Context 4} & \textbf{Models used} \\
\midrule
\endfirsthead
\multicolumn{6}{l}{\textit{Table S1 continued}} \\
\toprule
\textbf{Word} & \textbf{Context 1} & \textbf{Context 2} & \textbf{Context 3} & \textbf{Context 4} & \textbf{Models used} \\
\midrule
\endhead
\midrule
\multicolumn{6}{r}{\textit{Continued on next page}} \\
\endfoot
\bottomrule
\endlastfoot
bank & The bank of the river was muddy. & I need to go to the bank to deposit money. & Don't bank on getting help from them. &  & GPT-2, Llama, Qwen \\
letter & She wrote a letter to her friend overseas. & The letter A appears three times in this sentence. &  &  & GPT-2, Llama \\
right & Your answer is right. & Turn right at the next corner. &  &  & GPT-2, Llama \\
bat & The cave was full of bats hanging upside down. & He hit the ball with his bat. &  &  &  \\
band & The rock band played until midnight. & She used a band to tie her hair back. &  &  & GPT-2, Llama, Qwen \\
bear & The grizzly bear caught salmon in the stream. & She couldn't bear the weight of her grief. & That old apple tree still bears fruit every year. &  &  \\
crane & The crane waded through the shallow water. & The crane lifted the steel beam. & Crane your neck to see over the crowd. &  & GPT-2, Llama \\
ring & She wore a gold ring on her finger. & The children sat on the floor in a ring. & Please ring the doorbell when you arrive. &  & GPT-2, Llama, Qwen \\
palm & Tropical palm trees lined the beach. & He had the answer in the palm of his hand. &  &  & GPT-2, Llama, Qwen \\
spring & After a cold winter, flowers bloom in spring. & The mattress spring poked through the fabric. & They drank from the natural spring. & The cat was ready to spring on the mouse. & GPT-2, Llama, Qwen \\
pen & I need a pen for writing. & I need a pen for pigs. &  &  & GPT-2, Llama, Qwen \\
cast & They have a cast for acting. & They have a cast for healing. &  &  & GPT-2, Llama, Qwen \\
pitcher & The glass pitcher holds water. & The best pitcher throws fastballs. &  &  & GPT-2, Llama, Qwen \\
trunk & The car trunk holds luggage. & The tree trunk grows tall. &  &  & GPT-2, Llama, Qwen \\
bark & The dog bark scares people. & The tree bark feels rough. &  &  & GPT-2, Llama, Qwen \\
well & The water well is deep. & She sings well at concerts. &  &  & GPT-2, Llama, Qwen \\
book & Would you please book my reservation. & Please read my book now. &  &  & GPT-2, Llama, Qwen \\
light & The feather is very light and soft. & Please turn on the light and see. &  &  & GPT-2, Llama, Qwen \\
rock & The heavy rock fell down. & The loud rock played on. &  &  & GPT-2, Llama, Qwen \\
bow & The wooden bow shoots arrows. & The formal bow shows respect. & The bow of the ship cut through the waves. &  & GPT-2, Llama, Qwen \\
watch & I check my watch for time. & I will watch for danger. &  &  & GPT-2, Llama, Qwen \\
box & Put it in the box now. & He will box the champion. &  &  & GPT-2, Llama, Qwen \\
mine & This book is mine now. & This gold is from a mine now. &  &  & GPT-2, Llama, Qwen \\
fan & I am your fan forever. & I need a fan for cooling. &  &  & GPT-2, Llama, Qwen \\
fall & Don't fall down here. & We love fall colors here. &  &  & GPT-2, Llama, Qwen \\
stick & The wooden stick broke easily. & Can you stick around longer. &  &  & GPT-2, Llama, Qwen \\
rose & The rose smells beautiful. & He rose quickly upward. &  &  & GPT-2, Llama \\
saw & I saw him yesterday. & The saw cuts wood easily. &  &  & GPT-2, Llama \\
can & The soup can is empty. & I can help you tomorrow. &  &  & GPT-2, Llama \\
address & Please write your home address on the envelope. & The president will address the nation tonight. & We need to address this problem immediately. &  & GPT-2, Llama, Qwen \\
air & Open the window to let in some fresh air. & The pianist played a beautiful air from Mozart. & She has an air of confidence about her. &  & GPT-2, Llama, Qwen \\
arm & She broke her arm playing soccer. & The military decided to arm the border guards. & The research arm of the company is very innovative. &  & GPT-2, Llama, Qwen \\
bill & The duck's bill was bright orange. & Please pay the bill before you leave. & Congress will vote on the new bill tomorrow. &  & GPT-2, Llama, Qwen \\
bowl & Pour the cereal into a bowl. & He likes to bowl on Friday nights. & The team played at the super bowl. &  & GPT-2, Llama, Qwen \\
cabinet & Store the dishes in the kitchen cabinet. & The president met with his cabinet today. & She keeps files in a wooden cabinet. &  & GPT-2, Llama, Qwen \\
case & Pack your clothes in a suitcase. & The lawyer presented her case to the jury. & In this case, we should wait for more information. &  &  \\
change & Keep the change from the twenty dollar bill. & We need to change our approach to this problem. & Climate change affects everyone on the planet. &  & GPT-2, Llama, Qwen \\
charge & There is no charge for this service. & The cavalry began to charge across the field. & Make sure to charge your phone overnight. &  & GPT-2, Llama, Qwen \\
chest & He felt pain in his chest after running. & The pirates buried the treasure chest. & She keeps blankets in an old cedar chest. &  & GPT-2, Llama, Qwen \\
clip & Use a paper clip to hold these together. & Could you clip the article from the newspaper. & That was a funny video clip. &  & GPT-2, Llama, Qwen \\
club & She joined the chess club at school. & The caveman carried a wooden club. & They danced at the nightclub until dawn. &  &  \\
coil & The snake formed a tight coil. & Could you coil the rope neatly after use. & The electrical coil generates heat. &  & GPT-2, Llama, Qwen \\
count & You need to count to ten before you speak. & The old count lived in a grand castle. & Your opinion doesn't count in this decision. &  & GPT-2, Llama, Qwen \\
court & The trial will be held in court tomorrow. & They played basketball on the outdoor court. & The young man tried to court her affection. &  & GPT-2, Llama, Qwen \\
cover & Put a cover on the pot while it cooks. & This insurance policy will cover the damage. & The book has a beautiful cover design. &  & GPT-2, Llama, Qwen \\
current & The river has a strong current today. & What is the current temperature outside? & The electrical current flows through the wire. &  & GPT-2, Llama, Qwen \\
date & Write today's date at the top of the page. & Would you like to go on a date with me? & The date palm produces sweet fruit. &  & GPT-2, Llama, Qwen \\
die & All living things must die eventually. & Roll the die to see what number you get. & The factory uses a metal die for stamping. &  & GPT-2, Llama, Qwen \\
down & Walk down the stairs carefully. & The pillow is filled with goose down. & She felt down after hearing the bad news. &  & GPT-2, Llama, Qwen \\
drawer & Put the socks in the top drawer. & She is a talented drawer of portraits. & The antique drawer is made of oak. &  & GPT-2, Llama, Qwen \\
drill & Use a power drill to make the hole. & The soldiers performed a training drill. & The fabric has a tight drill weave. &  & GPT-2, Llama, Qwen \\
duck & A gray duck swam across the pond. & You should duck your head to avoid hitting the beam. & The fabric is called cotton duck. &  & GPT-2, Llama, Qwen \\
fair & That decision doesn't seem fair to me. & We went to the county fair last weekend. & She has fair skin that burns easily. &  & GPT-2, Llama, Qwen \\
fast & The cheetah can run very fast. & Muslims fast during the month of Ramadan. & Make sure the rope is tied fast. &  & GPT-2, Llama, Qwen \\
file & Put the documents in the file folder. & Use a metal file to smooth the rough edges. & People stood in a single file line. &  & GPT-2, Llama, Qwen \\
fine & She had to pay a parking fine. & The weather today is fine and sunny. & That is a fine piece of craftsmanship. &  & GPT-2, Llama, Qwen \\
firm & She works for a law firm downtown. & The mattress is too firm for my back. & He shook her hand with a firm grip. &  & GPT-2, Llama, Qwen \\
fit & These shoes fit me perfectly. & She stays fit by exercising daily. & He had a sudden fit of anger. &  & GPT-2, Llama, Qwen \\
fluke & The flatfish is also called a fluke. & Winning the lottery was a lucky fluke. & The anchor's fluke dug into the seafloor. &  &  \\
fly & Swat that fly before it lands on the food. & Birds often fly south for the winter. & His pants have a broken fly zipper. &  & GPT-2, Llama, Qwen \\
foot & She hurt her foot while hiking. & The measurement is exactly one foot tall. & Write your name at the foot of the page. &  & GPT-2, Llama, Qwen \\
grave & They placed flowers on her grave. & This is a grave matter requiring serious attention. & The accent grave appears in French words. &  & GPT-2, Llama, Qwen \\
ground & The ball fell to the ground. & She ground the coffee beans this morning. & Make sure the electrical outlet is properly ground. &  & GPT-2, Llama, Qwen \\
gum & The dentist examined each gum carefully. & Would you like a piece of chewing gum? & The eucalyptus gum tree is native to Australia. &  & GPT-2, Llama, Qwen \\
hamper & Put the dirty clothes in the laundry hamper. & Bad weather will hamper our travel plans. & She received a gift hamper full of treats. &  &  \\
hand & She held the baby in her hand. & Please hand me that book from the shelf. & The farm hand worked from dawn to dusk. &  & GPT-2, Llama, Qwen \\
head & She bumped her head on the low ceiling. & He will head the committee this year. & Pour the beer slowly to reduce the head. &  & GPT-2, Llama, Qwen \\
hide & The children love to hide during the game. & The leather comes from cattle hide. & He built a hunting hide in the forest. &  & GPT-2, Llama, Qwen \\
hit & Do not hit your brother with that toy. & That song was a huge hit last summer. & The website got a million hits today. &  &  \\
hold & Please hold my hand while we cross the street. & The ship's cargo is stored in the hold. & The wrestler used a tight hold. &  & GPT-2, Llama, Qwen \\
jam & She spread strawberry jam on her toast. & The photocopier has a paper jam again. & The musicians will jam together tonight. &  & GPT-2, Llama, Qwen \\
kind & What kind of music do you like? & She is a very kind and gentle person. & All people of every kind deserve respect. &  & GPT-2, Llama, Qwen \\
lap & The cat sat on her lap purring. & He completed another lap around the track. & The dog will lap water from the bowl. &  & GPT-2, Llama, Qwen \\
last & This is the last piece of cake. & These boots will last for many years. & The cobbler used a wooden shoe last. &  & GPT-2, Llama, Qwen \\
lead & The pipes are made of lead metal. & She will lead the team to victory. & Walk the dog on a short lead. &  & GPT-2, Llama, Qwen \\
lean & Don't lean against that wet wall. & He is tall and lean with no extra weight. & Choose lean meat for a healthier diet. &  & GPT-2, Llama, Qwen \\
leave & We should leave before it gets dark. & He took a week of sick leave from work. & The trees shed their leaves in autumn. &  &  \\
left & Be sure to turn left at the traffic light. & She had left the party early last night. & There is no food left in the refrigerator. &  & GPT-2, Llama, Qwen \\
lie & Please lie down and rest for a while. & Don't lie to me about what happened. & The town lies in a beautiful valley. &  &  \\
line & Draw a straight line with your ruler. & Stand in line and wait your turn. & He forgot his lines during the play. &  &  \\
long & The river is very long and winding. & I long to see my family again. & It won't be long before dinner is ready. &  & GPT-2, Llama, Qwen \\
lot & She has a lot of books on that shelf. & They bought a building lot downtown. & He drew the short lot and had to go first. &  & GPT-2, Llama, Qwen \\
match & Light the candle with a wooden match. & The red dress will match your shoes perfectly. & The tennis match lasted three hours. &  & GPT-2, Llama, Qwen \\
mean & Calculate the mean of these five numbers. & Don't be mean to your little sister. & What does this word mean in English? &  & GPT-2, Llama, Qwen \\
mole & The doctor removed a mole from his back. & A mole dug tunnels under the lawn. & She works as a mole for the intelligence agency. &  & GPT-2, Llama, Qwen \\
mount & They plan to mount an expedition next spring. & He said Mount Everest is the highest peak on Earth. & The cavalry officer sat on his mount. &  &  \\
nail & Hammer the nail into the wood. & She painted each nail bright red. & You should nail the presentation tomorrow &  & GPT-2, Llama, Qwen \\
net & The fisherman cast his net into the sea. & The company's net profit increased this quarter. & Hit the tennis ball over the net. &  & GPT-2, Llama, Qwen \\
note & She left a note on the kitchen table. & The singer hit the high note perfectly. & Please note the change in schedule. &  & GPT-2, Llama, Qwen \\
novel & She is reading a mystery novel. & That is a novel approach to the problem. & The invention was truly novel and innovative. &  & GPT-2, Llama, Qwen \\
order & The files are arranged in alphabetical order. & I would like to order the salmon for dinner. & The judge issued a court order. &  & GPT-2, Llama, Qwen \\
organ & The heart is a vital organ. & She plays the organ at church on Sundays. & The newspaper is the official organ of the party. &  & GPT-2, Llama, Qwen \\
part & This is the best part of the movie. & They decided to part ways after graduation. & She will play the part of Juliet. &  & GPT-2, Llama, Qwen \\
pass & Could you pass the salt and pepper. & You need a pass to enter the building. & The mountain pass was covered with snow. &  & GPT-2, Llama, Qwen \\
patient & The doctor examined the patient carefully. & Please be patient while we process your request. & She has a patient and calm personality. &  & GPT-2, Llama, Qwen \\
pick & Pick an apple from the tree. & He used a pick to break up the ice. & That was the first pick in the draft. &  & GPT-2, Llama, Qwen \\
pine & The pine tree stays green all winter. & She continued to pine for her lost love. & The furniture is made of solid pine wood. &  & GPT-2, Llama, Qwen \\
pitch & The baseball player has a great pitch. & They will pitch their tent near the river. & The roof has a steep pitch. & The ship rolled in the pitch and roll of the waves. & GPT-2, Llama, Qwen \\
place & This is a beautiful place to visit. & Would you place the books on the table. & She finished in third place in the race. &  & GPT-2, Llama, Qwen \\
plane & The plane took off on schedule. & Use a plane to smooth the wooden surface. & The geometry problem involves a flat plane. &  & GPT-2, Llama, Qwen \\
plant & Water the plant twice a week. & The factory plant operates around the clock. & Farmers plant the seeds in early spring. &  & GPT-2, Llama, Qwen \\
plate & Put the food on a clean plate. & The car has a California license plate. & The Earth's tectonic plates move slowly. &  &  \\
play & Children love to play in the park. & We saw an excellent play at the theater. & There is some play in the steering wheel. &  & GPT-2, Llama, Qwen \\
plot & The novel has a complex plot. & They bought a plot of land for building. & The conspirators plot against the king. &  & GPT-2, Llama, Qwen \\
point & The pencil has a sharp point. & What is the point of this exercise? & Can you point to the correct answer. & The decimal point separates whole numbers from fractions. & GPT-2, Llama, Qwen \\
pole & Hang the flag on the tall pole. & The North Pole is covered in ice. & She used a pole to reach the high shelf. &  &  \\
pool & The children swam in the pool all afternoon. & Let's pool our resources to buy the gift. & He plays pool at the bar on weekends. &  & GPT-2, Llama, Qwen \\
pound & The recipe calls for one pound of flour. & Use a hammer to pound the nail. & The British pound is the currency of England. &  & GPT-2, Llama, Qwen \\
press & Press the button to start the machine. & The printing press revolutionized communication. & The press reported on the scandal extensively. &  & GPT-2, Llama \\
punch & He threw a punch at his opponent. & Would you like some fruit punch? & Use the punch to make holes in the leather. &  & GPT-2, Llama, Qwen \\
quarry & Marble is extracted from a stone quarry. & The predator began to stalk its quarry across the plain. & The hunter lost sight of his quarry. &  & GPT-2, Llama, Qwen \\
rail & Hold onto the rail as you go downstairs. & The train runs on steel rails. & He continued to rail against the decision. &  &  \\
range & The mountain range stretches for miles. & Cook the meat on the gas range. & The store offers a wide range of products. &  & GPT-2, Llama, Qwen \\
rank & She holds the rank of captain. & These are rank weeds with a foul smell. & Next rank the items from most to least important. &  & GPT-2, Llama, Qwen \\
rare & That is a rare and valuable stamp. & I like my steak cooked rare. & It is rare to see snow in this region. &  & GPT-2, Llama, Qwen \\
rate & What is the interest rate on the loan? & How would you rate this restaurant? & The heart rate increased during exercise. &  & GPT-2, Llama, Qwen \\
rest & Sit down and rest for a few minutes. & The rest of the group will arrive later. & The book rests on the table. &  &  \\
root & The tree has a deep root in the ground. & The fans will root for their team. & Let's get to the root of the problem. &  & GPT-2, Llama, Qwen \\
row & They sat in the front row at the theater. & She will row the boat across the lake. & The houses are built in a neat row. &  & GPT-2, Llama, Qwen \\
ruler & Use a ruler to measure the length. & The ruler governed the kingdom wisely. & Draw a line with a straight ruler. &  & GPT-2, Llama, Qwen \\
run & She can run five miles without stopping. & The play will run for three months. & There is a run in her stocking. &  & GPT-2, Llama, Qwen \\
sage & The old man was known as a wise sage. & Add fresh sage to the stuffing. & She offered sage advice to the young couple. &  & GPT-2, Llama, Qwen \\
scale & The fish has shiny silver scales. & Weigh yourself on the bathroom scale. & They plan to scale the mountain next week. &  &  \\
season & Autumn is my favorite season of the year. & You must season the soup with salt and pepper. & The wood needs time to season properly. &  & GPT-2, Llama, Qwen \\
second & Wait just a second before you leave. & She finished in second place in the race. & I would like a second helping of dessert. &  & GPT-2, Llama, Qwen \\
sentence & Write a complete sentence with proper punctuation. & The judge will sentence the defendant tomorrow. & Read the first sentence of each paragraph. &  & GPT-2, Llama, Qwen \\
shed & Store the tools in the garden shed. & The snake will shed its skin soon. & She shed tears of joy at the wedding. &  & GPT-2, Llama, Qwen \\
shop & The bakery shop opens at six o'clock. & She loves to shop for new clothes. & He runs a small repair shop downtown. &  & GPT-2, Llama, Qwen \\
side & Stand on the left side of the room. & I will side with you in this argument. & Order a side of vegetables with your meal. &  & GPT-2, Llama, Qwen \\
sign & Read the traffic sign carefully. & You need to sign your name at the bottom. & Dark clouds are a sign of approaching rain. &  & GPT-2, Llama, Qwen \\
sink & Wash the dishes in the kitchen sink. & The ship began to sink beneath the waves. & Don't let yourself sink into depression. &  & GPT-2, Llama, Qwen \\
skip & The children skip rope in the playground. & Please don't skip any pages in the book. & Throw the stone so it will skip across the water. &  & GPT-2, Llama, Qwen \\
slip & Be careful not to slip on the wet floor. & She wrote a note on a slip of paper. & The boat is stored at the marina slip. &  & GPT-2, Llama, Qwen \\
slug & A slug crawled across the garden path. & He took a slug of whiskey from the bottle. & The boxer landed a hard slug to the jaw. &  & GPT-2, Llama, Qwen \\
spell & Can you spell your name for me? & The witch cast a magic spell. & We had a spell of cold weather last week. &  & GPT-2, Llama, Qwen \\
spoke & She spoke clearly during the presentation. & The bicycle wheel has thirty spokes. & One spoke on the wheel is broken. &  &  \\
spot & There is a spot of ink on your shirt. & This is the perfect spot for a picnic. & Can you spot the difference between these pictures? &  & GPT-2, Llama, Qwen \\
squash & Grow pumpkins and squash in the garden. & Don't squash the bread in the bag. & They play squash ats the gym twice a week. &  & GPT-2, Llama, Qwen \\
stalk & Each celery stalk should be washed carefully. & The predator began to stalk its prey. & He would stalk her on social media constantly. &  & GPT-2, Llama, Qwen \\
stamp & Put a stamp on the envelope before mailing. & She will stamp her foot when she's angry. & The document needs an official stamp. &  & GPT-2, Llama, Qwen \\
staple & Use a staple to fasten the pages together. & Rice is a staple food in many cultures. & Cotton was the staple crop of the region. &  & GPT-2, Llama, Qwen \\
steep & The hill is too steep to climb easily. & Steep the tea bag in hot water. & That price seems rather steep to me. &  & GPT-2, Llama, Qwen \\
stern & The teacher gave them a stern warning. & The stern of the boat faces backward. & He has a stern and serious expression. &  & GPT-2, Llama, Qwen \\
strike & The workers will strike for better wages. & You can trike the match to light the candle. & The clock will strike twelve at midnight. &  & GPT-2, Llama, Qwen \\
suit & He wore a dark blue business suit. & The lawyer filed a civil suit. & Pick a card from any suit in the deck. &  & GPT-2, Llama, Qwen \\
swallow & A swallow built its nest under the eaves. & Chew your food well before you swallow. & That was a bitter pill to swallow. &  & GPT-2, Llama, Qwen \\
table & Set the dishes on the dining room table. & They decided to table the discussion until later. & Study the table of contents carefully. &  & GPT-2, Llama, Qwen \\
tap & Turn on the tap to get some water. & She will tap her fingers when she's nervous. & The dancer performed an amazing tap routine. &  & GPT-2, Llama, Qwen \\
temple & The ancient temple was built of stone. & He felt a throbbing pain in his temple. & Pray at the temple every morning. &  & GPT-2, Llama, Qwen \\
tender & The meat is very tender and juicy. & She has a tender and caring nature. & Submit your tender for the construction project. &  & GPT-2, Llama, Qwen \\
tick & The clock makes a loud tick every second. & Put a tick mark next to correct answers. & Check yourself for ticks after hiking. &  &  \\
till & Wait till tomorrow before you decide. & Count the money in the cash till. & Farmers till the soil in early spring. &  & GPT-2, Llama, Qwen \\
tip & The tip of the pencil broke off. & Leave a generous tip for the waiter. & I'll give you a helpful tip about studying. &  & GPT-2, Llama, Qwen \\
toast & Butter the toast while it's still warm. & Let's make a toast to the happy couple. & The bread is in the toaster for toast. &  & GPT-2 \\
track & The train runs on the railroad track. & Keep track of your expenses this month. & They found animal tracks in the snow. &  &  \\
train & The train departs at three o'clock. & She will train for the marathon all year. & The wedding dress has a long train. &  & GPT-2, Llama, Qwen \\
trip & They took a trip to Europe last summer. & Don't trip over that loose carpet. & The circuit breaker will trip if overloaded. &  & GPT-2, Llama, Qwen \\
wave & The wave crashed against the rocks. & Wave goodbye to your friends. & She has beautiful waves in her hair. &  &  \\
will & I will see you tomorrow morning. & She has a strong will to succeed. & Read the terms in his last will. &  & GPT-2, Llama, Qwen \\
yard & The children played in the back yard. & The fabric is sold by the yard. & The boat is thirty yards from shore. &  &  \\
yield & The farm will yield a good harvest. & Drivers must yield to traffic from the right. & The investment has a high yield. &  & GPT-2, Llama, Qwen \\
wind & The wind blew strongly across the plains. & Please wind the string around the spool tightly. & The river has many twists and turns as it winds through the valley. &  &  \\
tear & A single tear rolled down her cheek. & Be careful not to tear the delicate paper. & There is a large tear in his jacket. &  & GPT-2, Llama, Qwen \\
live & Many species live in the rainforest. & The concert will be broadcast live tonight. & Handle the live electrical wire with care. &  & GPT-2, Llama, Qwen \\
close & Make sure to close the door when you leave. & The store is close to my house. & That was a close call with the car. &  & GPT-2, Llama, Qwen \\
minute & Wait just one minute before you go. & The details are so minute they're hard to see. & Write down the minutes from the meeting. &  &  \\
wound & The soldier had a deep wound in his leg. & She wound the clock every morning. & The path wound through the forest. &  & GPT-2, Llama, Qwen \\
sow & The farmer will sow seeds in spring. & The sow gave birth to ten piglets. & We need to sow discord among our enemies. &  & GPT-2, Llama, Qwen \\
does & She does her homework every evening. & Several does grazed in the meadow. & He does what he thinks is right. &  & GPT-2, Llama \\
entrance & The entrance to the building is on the left. & The magician will entrance the audience with his tricks. & Use the main entrance for deliveries. &  & GPT-2, Llama, Qwen \\
present & I bought her a birthday present. & Please be present at the meeting tomorrow. & The lawyer will present evidence to the court. &  & GPT-2, Llama, Qwen \\
refuse & The garbage refuse smells terrible. & I refuse to accept these terms. & They refuse to negotiate further. &  & GPT-2, Llama \\
subject & Math is my favorite subject in school. & The king will subject his enemies to harsh treatment. & All citizens are subject to the law. &  & GPT-2, Llama, Qwen \\
object & What is that strange object in the sky? & I object to this unfair proposal. & The object of the game is to score points. &  & GPT-2, Llama \\
produce & Buy fresh produce at the farmers market. & The factory can produce one thousand units daily. & This region will produce excellent wine. &  & GPT-2, Llama, Qwen \\
record & Keep a written record of all expenses. & Make sure to record the meeting for later review. & She broke the world record for speed. &  & GPT-2, Llama, Qwen \\
permit & You need a permit to park here. & The rules do not permit smoking inside. & Weather conditions will permit us to travel. &  & GPT-2, Llama, Qwen \\
project & The science project is due next Friday. & They project sales will increase next year. & The balcony will project out over the street. &  & GPT-2, Llama, Qwen \\
rebel & The rebel forces attacked the capital. & Teenagers often rebel against their parents. & She has always been a rebel at heart. &  & GPT-2, Llama, Qwen \\
contract & Sign the contract before you start work. & Metal will contract when it gets cold. & The business has a contract with the government. &  & GPT-2, Llama, Qwen \\
content & She felt content with her simple life. & The content of the book is fascinating. & Please review the content before publishing. &  & GPT-2, Llama, Qwen \\
desert & The Sahara desert is extremely hot and dry. & A soldier should never desert his post. & They were stranded on a desert island. &  & GPT-2, Llama, Qwen \\
bass & He caught a large bass in the lake. & She sings bass in the choir. & The bass player sets the rhythm. &  & GPT-2, Llama, Qwen \\
invalid & Your password is invalid and must be reset. & The invalid requires constant medical care. & This ticket is invalid after six months. &  & GPT-2, Llama, Qwen \\
resume & Submit your resume with the application. & We will resume the meeting after lunch. &  &  & GPT-2, Llama, Qwen \\
sewer & The sewer system needs repair. & She is an expert sewer of fine garments. & Raw sewage flows through the sewer pipes. &  & GPT-2, Llama, Qwen \\
compound & Water is a chemical compound of hydrogen and oxygen. & The refugees lived in a fenced compound. & This will compound the existing problems. &  & GPT-2, Llama, Qwen \\
attribute & Patience is an important attribute. & We attribute our success to hard work. & The painting has been attribute to Rembrandt. &  & GPT-2, Llama, Qwen \\
console & The gaming console is very popular. & She tried to console her crying friend. & Press the button on the console. &  & GPT-2, Llama, Qwen \\
duplicate & Make a duplicate copy of the key. & We need to duplicate these results. & This is a duplicate of the original document. &  & GPT-2, Llama, Qwen \\
\end{longtable}
}
\clearpage
\noindent{\bfseries Table S2}\par
\noindent Logit lens concept-drift polyseme stimuli: candidate pool of concept-drift pairs and sentence contexts, prior to per-model validation\par\smallskip
{\fontsize{6.7}{8.0}\selectfont
\begin{longtable}{@{}>{\raggedright\arraybackslash}p{0.09\linewidth}>{\raggedright\arraybackslash}p{0.215\linewidth}>{\raggedright\arraybackslash}p{0.215\linewidth}>{\raggedright\arraybackslash}p{0.215\linewidth}>{\raggedright\arraybackslash}p{0.16\linewidth}@{}}
\toprule
\textbf{Word} & \textbf{Context 1} & \textbf{Context 2} & \textbf{Context 3} & \textbf{Models used} \\
\midrule
\endfirsthead
\multicolumn{5}{l}{\textit{Table S2 continued}} \\
\toprule
\textbf{Word} & \textbf{Context 1} & \textbf{Context 2} & \textbf{Context 3} & \textbf{Models used} \\
\midrule
\endhead
\midrule
\multicolumn{5}{r}{\textit{Continued on next page}} \\
\endfoot
\bottomrule
\endlastfoot
mouse & The cat chased the mouse across the kitchen floor. & I need to replace the batteries in my wireless mouse. &  & GPT-2, Llama, Qwen \\
tablet & Take two tablets with water after each meal. & She downloaded the reading app on her tablet. & The ancient tablet contained hieroglyphic inscriptions. &  \\
stream & The mountain stream flowed gently through the valley. & I love to stream movies on my laptop at night. &  & GPT-2, Llama, Qwen \\
drive & She learned to drive when she turned sixteen. & Please save the document to your Google drive. & Her drive to succeed motivated everyone around her. & GPT-2, Llama, Qwen \\
cloud & The dark cloud promised rain later today. & All our photos are backed up in the cloud. &  & GPT-2, Llama, Qwen \\
viral & The viral infection spread rapidly through the school. & Her dance video went viral on social media overnight. &  & GPT-2, Llama \\
tweet & The robin's sweet tweet woke me at dawn. & Her tweet about the conference got a thousand likes. &  & GPT-2, Llama \\
ghost & The old mansion was rumored to have a ghost. & He completely ghosted her after their third date. & The ghost writer helped him finish his memoir. & GPT-2, Llama, Qwen \\
smart & She was the smartest student in her class. & I bought a new smart TV for the living room. &  & Llama, Qwen \\
gig & The band played a gig at the local venue. & He works gig jobs to pay for college tuition. &  & GPT-2, Llama, Qwen \\
slide & The children played on the playground slide. & He tried to slide into her DMs last night. & Please advance to the next slide in the presentation. & GPT-2, Llama, Qwen \\
flex & The gymnast could flex her back remarkably. & He loves to flex his expensive watch online. &  & GPT-2, Llama, Qwen \\
cancel & We had to cancel dinner due to the storm. & The internet tried to cancel the celebrity yesterday. &  & GPT-2, Llama, Qwen \\
fire & The fire crackled warmly in the fireplace. & That new song is absolutely fire! &  & GPT-2, Llama \\
sick & The child stayed home because he was sick. & Those skateboard tricks were totally sick! &  & GPT-2, Llama, Qwen \\
troll & The fairy tale featured a troll under the bridge. & Don't feed the troll in the comment section. &  & GPT-2, Llama, Qwen \\
spam & We had fried spam and eggs for breakfast. & My email inbox is full of spam messages. &  & GPT-2, Llama, Qwen \\
cookie & She baked chocolate chip cookies for dessert. & The website uses cookies to track user behavior. &  & GPT-2, Llama, Qwen \\
bug & A lady bug crawled across the garden leaf. & The software has a bug that crashes the program. & They decided to bug the suspect's phone line. & GPT-2, Llama, Qwen \\
crash & The car crash blocked traffic for hours. & My computer will crash if I open too many programs. & I need to crash at your place tonight. & GPT-2, Llama, Qwen \\
hack & He used an axe to hack through the thick wood. & Criminals tried to hack into the bank's database. & That's a useful life hack for saving time. & GPT-2, Llama, Qwen \\
surf & We love to surf the big waves in California. & I like to surf the internet during lunch breaks. &  & GPT-2, Llama, Qwen \\
web & The spider spun an intricate web in the corner. & I found the answer by searching the web. &  & GPT-2, Llama, Qwen \\
host & The gracious host welcomed guests into her home. & The web server host crashed during peak traffic hours. &  & GPT-2, Llama, Qwen \\
server & The dapper server brought our meals promptly to the table. & The web server handles thousands of requests daily. &  & GPT-2, Llama, Qwen \\
platform & The train arrived at platform nine exactly on time. & YouTube is a popular video sharing platform. &  & GPT-2, Llama, Qwen \\
profile & The artist drew her profile from the side view. & I updated my LinkedIn profile with new skills. &  & GPT-2, Llama, Qwen \\
feed & The farmer went to feed the hungry chickens. & I scrolled through my social media feed. &  & GPT-2, Llama, Qwen \\
friend & She has been my closest friend since childhood. & Can you friend me on Facebook when you get home. &  & GPT-2, Llama, Qwen \\
post & The fence post was made of sturdy oak wood. & Her Instagram post received hundreds of comments. & Please post this letter when you go to town. & GPT-2, Llama, Qwen \\
tag & She removed the price tag from her new dress. & Please tag me in the photo from last night. & The children played tag in the playground. & GPT-2, Llama, Qwen \\
link & The chain had a broken link near the middle. & Click this link to visit our company website. &  & GPT-2, Llama, Qwen \\
thread & The seamstress used gold thread for the embroidery. & The forum thread discussed politics for hours. &  & GPT-2, Llama, Qwen \\
handle & The door handle was made of polished brass. & My Twitter handle reflects my interests perfectly. &  & GPT-2, Llama, Qwen \\
window & She opened the bedroom window to let in air. & Please close that browser window when you're finished. &  & GPT-2, Llama, Qwen \\
file & The carpenter used a metal file to smooth edges. & Please save this document as a PDF file. & The detective studied each file in the case. & GPT-2, Llama, Qwen \\
menu & The restaurant menu featured authentic Italian dishes. & Right-click to open the context menu options. &  & GPT-2, Llama, Qwen \\
page & Turn to page fifty-seven in your textbook. & The web page loaded slowly on my phone. &  & GPT-2, Llama, Qwen \\
site & The construction site was busy with workers. & I visit that news site every morning. &  & GPT-2, Llama, Qwen \\
address & Please write your home address on this form. & Type the website address into your browser. & The president will address the nation tonight. & GPT-2, Llama, Qwen \\
domain & The king ruled over his vast domain wisely. & I registered a new domain name for my business. &  & GPT-2, Llama, Qwen \\
boot & She wore a leather boot on her injured foot. & Please boot the computer in safe mode. &  & GPT-2, Llama \\
cache & Pirates hid their treasure cache on the island. & Clear your browser cache to fix the problem. &  & GPT-2, Llama, Qwen \\
chip & He found a chip of wood in his finger. & The computer chip processes billions of calculations. &  & GPT-2, Llama, Qwen \\
code & The spy used a secret code for messages. & She writes code in Python every day. &  & GPT-2, Llama, Qwen \\
desktop & Clear the desktop of your desk please. & My computer desktop is cluttered with icons. &  & GPT-2, Llama, Qwen \\
folder & Put the papers in that manila folder. & Create a new folder for these files. &  & GPT-2, Llama, Qwen \\
icon & The religious icon hung above the altar. & Click the icon to open the program. &  & GPT-2, Llama, Qwen \\
keyboard & The piano keyboard has eighty-eight keys. & My computer keyboard needs new batteries. &  & GPT-2, Llama, Qwen \\
memory & Her memory of childhood was very vivid. & The computer needs more memory for this task. &  & GPT-2, Llama \\
net & The fisherman cast his net into the water. & I found the information on the internet. &  &  \\
network & She built a network of professional contacts. & The computer network went down this morning. &  & GPT-2, Llama, Qwen \\
notebook & She wrote notes in her leather notebook. & I bought a new notebook computer yesterday. &  & GPT-2, Llama, Qwen \\
paste & Apply the paste evenly to the wall. & Just copy and paste the text here. &  & GPT-2, Llama, Qwen \\
portal & The stone portal led into the castle. & Log into the student portal for grades. &  & GPT-2, Llama, Qwen \\
program & The evening program includes three performances. & She learned to program computers at university. &  & GPT-2, Llama, Qwen \\
screen & The privacy screen blocked prying eyes. & My computer screen is too bright. &  & GPT-2, Llama, Qwen \\
scroll & The ancient scroll contained sacred texts. & You need to scroll down to read more. &  & GPT-2, Llama, Qwen \\
terminal & The bus terminal was crowded with travelers. & Open the computer terminal to run commands. &  & GPT-2, Llama, Qwen \\
toolbar & The carpenter organized tools on his toolbar. & Click the icon on the toolbar. &  & GPT-2, Llama, Qwen \\
traffic & Heavy traffic delayed our arrival significantly. & Website traffic increased after the promotion. &  & GPT-2, Llama \\
virus & The airborne virus caused a terrible flu outbreak. & A computer virus infected the entire network. &  & GPT-2, Llama, Qwen \\
block & The child played with a wooden block. & I had to block that annoying user online. &  & GPT-2, Llama, Qwen \\
filter & Pour the coffee through a paper filter. & She applied a filter to her photo. & The web filter blocks inappropriate content automatically. & GPT-2, Llama, Qwen \\
follow & Would you follow me to the conference room. & Thanks, I'll follow your account on social media. &  & GPT-2, Llama, Qwen \\
footprint & We found a bear footprint in the mud. & Your digital footprint reveals browsing habits. &  & GPT-2, Llama, Qwen \\
ping & The sonar sends a ping through water. & Let me ping you when I arrive. &  & GPT-2, Llama, Qwen \\
share & Everyone should share their toys fairly. & Please share this article on social media. &  & GPT-2, Llama \\
status & Her social status improved after the promotion. & Update your status to show availability. &  & GPT-2, Llama, Qwen \\
timeline & The historical timeline spans three centuries. & Check my timeline for recent updates. &  & GPT-2, Llama, Qwen \\
wall & They painted the wall bright yellow. & Post it on my wall for everyone to see. &  & GPT-2, Llama, Qwen \\
cool & The evening air feels pleasantly cool tonight. & That new phone is really cool. &  & GPT-2, Llama, Qwen \\
lit & She took a lighter and lit the candles for the ceremony. & That party was absolutely lit last night. &  & GPT-2, Llama, Qwen \\
crush & Please don't crush the delicate flowers please. & She has a crush on her coworker. &  & GPT-2, Llama, Qwen \\
epic & Homer wrote an epic poem called the Iliad. & That skateboard trick was totally epic. &  & GPT-2, Llama, Qwen \\
savage & The large and savage beast roamed the jungle freely. & That comeback was absolutely savage honestly. &  & GPT-2, Llama, Qwen \\
ship & The cargo ship sailed across the ocean. & I totally ship those two characters together. &  & GPT-2, Llama, Qwen \\
tea & She prefers green tea to black tea. & Spill the tea about what happened. &  & GPT-2, Llama, Qwen \\
extra & Would you like extra cheese on that? & She's being so extra about everything today. &  & GPT-2, Llama, Qwen \\
shook & She shook the bottle before opening it. & I was shook by the surprise ending. &  & GPT-2, Llama \\
shade & We sat in the shade of the oak tree. & She threw shade at her rival subtly. &  & GPT-2, Llama, Qwen \\
trash & Take the trash out to the curb. & Critics will trash the movie unfairly. &  & GPT-2, Llama, Qwen \\
vibe & The speaker's vibration created a strange vibe. & This place has a really chill vibe. &  & GPT-2, Llama, Qwen \\
mood & Her cheerful mood lifted everyone's spirits. & This rainy weather is such a mood. &  & GPT-2, Llama, Qwen \\
cap & He wore a baseball cap backwards. & That story sounds like complete cap honestly. &  & GPT-2, Llama, Qwen \\
slick & The road was slick with ice this morning. & That was a slick move you made. &  & GPT-2, Llama, Qwen \\
sketch & The artist drew a quick pencil sketch. & That guy seems really sketch to me. &  & GPT-2, Llama, Qwen \\
wild & The zoo has wild animals from Africa. & That party was absolutely wild last night. &  & GPT-2, Llama, Qwen \\
fresh & We buy fresh vegetables from the market daily. & Those new kicks are really fresh. &  & GPT-2, Llama, Qwen \\
tight & The jar lid was screwed on tight. & We're really tight as best friends. &  & GPT-2, Llama, Qwen \\
bomb & Fortunately the bomb did not explode there. & This new restaurant is the bomb seriously. &  & GPT-2, Llama, Qwen \\
beat & Don't beat the drum too loudly please. & Her makeup look is absolutely beat today. &  & GPT-2, Llama, Qwen \\
bounce & The ball will bounce very high here. & Let's bounce from this party early tonight. &  & GPT-2, Llama, Qwen \\
flame & The candle flame flickered in the breeze. & That post started a flame war. &  & GPT-2, Llama, Qwen \\
snack & She ate a healthy snack between meals. & He's such a snack in that outfit. &  & GPT-2, Llama, Qwen \\
finesse & The jeweler crafted the ring with finesse. & You should finesse that phone number from her &  &  \\
mad & She was mad about missing her flight. & That's a mad amount of homework tonight. &  & GPT-2, Llama, Qwen \\
\end{longtable}
}
\clearpage
\noindent{\bfseries Table S3}\par
\noindent Activation patching homonym stimuli: position-matched candidate pool of homonym pairs, prior to per-model validation\par\smallskip
{\fontsize{6.7}{8.0}\selectfont
\begin{longtable}{@{}>{\raggedright\arraybackslash}p{0.10\linewidth}>{\raggedright\arraybackslash}p{0.32\linewidth}>{\raggedright\arraybackslash}p{0.32\linewidth}>{\raggedright\arraybackslash}p{0.18\linewidth}@{}}
\toprule
\textbf{Word} & \textbf{Context 1} & \textbf{Context 2} & \textbf{Models used} \\
\midrule
\endfirsthead
\multicolumn{4}{l}{\textit{Table S3 continued}} \\
\toprule
\textbf{Word} & \textbf{Context 1} & \textbf{Context 2} & \textbf{Models used} \\
\midrule
\endhead
\midrule
\multicolumn{4}{r}{\textit{Continued on next page}} \\
\endfoot
\bottomrule
\endlastfoot
air & The fresh air smells very clean & The classical air plays beautifully & GPT-2, Llama, Qwen \\
arm & She broke her arm playing soccer & They decided to arm the guards & GPT-2, Llama, Qwen \\
attribute & Her best attribute is patience & We can attribute success here & GPT-2, Llama, Qwen \\
band & The rock band plays music loudly & The rubber band snaps back quickly & GPT-2, Llama \\
bank & The steep bank holds the water & The local bank holds my money & GPT-2, Llama \\
bark & The loud bark scares small children & The rough bark protects the tree &  \\
bass & They caught a bass yesterday & They sang the bass part well & GPT-2 \\
bat & The brown bat hangs upside down & The wooden bat hits balls hard &  \\
bear & The big brown bear catches salmon well & She could not bear the heavy weight & GPT-2, Llama, Qwen \\
bill & The duck's bill is bright orange & Please pay the bill before leaving & GPT-2, Llama, Qwen \\
book & Please read my book about science & Please help me book a flight & GPT-2, Llama, Qwen \\
bow & The wooden bow shoots arrows far & The formal bow shows deep respect &  \\
bowl & The ceramic bowl holds fresh fruit & They often bowl on Friday nights & GPT-2, Llama, Qwen \\
box & The cardboard box holds old books & Professional fighters box in the ring & GPT-2, Llama, Qwen \\
cabinet & The kitchen cabinet stores clean dishes & The presidential cabinet meets today & Qwen \\
can & The metal can holds tomato soup & Yes I can help you now & GPT-2, Llama, Qwen \\
case & The leather case protects my camera & The legal case goes to trial &  \\
cast & The talented cast performs the play & The plaster cast protects broken bones & Llama, Qwen \\
change & Keep the change from the purchase & We must change our approach quickly & GPT-2, Llama, Qwen \\
charge & There's no charge for this service & You must now charge your phone & GPT-2, Llama, Qwen \\
chest & He feels a chest pain often & The old wooden chest holds treasure & GPT-2, Llama, Qwen \\
clip & The paper clip holds sheets together & Now please clip those hedges carefully & GPT-2, Llama, Qwen \\
close & You should now close the door & The store is close to house & GPT-2, Llama, Qwen \\
club & The chess club meets every Tuesday & The wooden club serves as weapon & Qwen \\
coil & The snake formed a tight coil & That copper wire has a coil & Llama, Qwen \\
compound & This creates a compound mixture & These errors will compound problems & GPT-2, Llama, Qwen \\
console & He wanted a console for his birthday & She aimed to console her friend & GPT-2, Llama, Qwen \\
content & She felt content with her life & The book content is very fascinating & GPT-2, Llama, Qwen \\
contract & Sign the new contract before you start & The metal can contract when cold & GPT-2, Qwen \\
count & Can you count to ten slowly & The old count lived in castle & GPT-2, Llama, Qwen \\
court & The legal court begins today & The tennis court needs repairs & Llama, Qwen \\
cover & The beautiful book cover is torn & This policy will cover the damage & GPT-2, Llama, Qwen \\
current & The strong river current flows & What is the current temperature & GPT-2, Llama, Qwen \\
date & Please write the exact date here & Would you like a date tonight & GPT-2, Llama, Qwen \\
die & All plants must die eventually & Now roll the die carefully & GPT-2, Qwen \\
does & She does her homework every evening & Several does grazed in the meadow & Qwen \\
drawer & The wooden drawer holds items & The portrait drawer works here & Llama \\
drill & The power drill makes clean holes & The fire drill happens every month &  \\
duplicate & We make a duplicate copy of the key & We need to duplicate these results & GPT-2, Llama, Qwen \\
fair & This seems fair to me & The county fair has rides & GPT-2, Llama, Qwen \\
fall & Don't fall down the steep stairs & The beautiful fall colors trees red & GPT-2, Llama, Qwen \\
fan & She is your best fan now & Can you turn the fan on & GPT-2, Llama, Qwen \\
file & The manila file holds papers & Use a metal file carefully & GPT-2, Llama, Qwen \\
fine & She received the fine yesterday & The weather looks fine today &  \\
fit & These shoes fit me very well & She stays fit through exercise & GPT-2, Llama, Qwen \\
fly & The annoying fly lands on food & Most birds fly south for winter & GPT-2, Llama, Qwen \\
grave & They visited her grave with flowers & This is a grave matter indeed & GPT-2, Llama, Qwen \\
hand & write with either hand to show me & Could you please hand me that & GPT-2, Llama, Qwen \\
head & She hit her head on the door & She will soon head to new cinema & GPT-2, Llama, Qwen \\
hide & They decided to hide & The armor is hide & GPT-2, Llama, Qwen \\
hit & The boxer landed a hard hit & The new movie was a big hit &  \\
invalid & Your old password is now invalid & The nurse cared for the invalid & Qwen \\
kind & He is a very kind and thoughtful person & What is your favorite kind of music & Llama, Qwen \\
last & This is the last piece remaining & These boots will last many years & GPT-2, Llama, Qwen \\
lean & You should not lean on that wall & The recipe requires lean ground beef & GPT-2, Llama, Qwen \\
leave & I must now leave for my appointment & She took a leave of absence from work & GPT-2, Llama, Qwen \\
left & Please turn left at the light & She had left the party early & GPT-2, Llama, Qwen \\
letter & She wrote a letter to friends & She traced the letter very carefully & Qwen \\
lie & You should not lie to your parents & He wanted to lie on the soft bed &  \\
light & The bright light illuminates the room & The small light package arrives soon & Llama, Qwen \\
line & Draw a straight line with ruler & Please stand in line and wait & GPT-2, Llama, Qwen \\
lot & There are a lot of cars here & They bought a lot to build on & Llama, Qwen \\
mine & That red car is definitely mine & The workers entered the coal mine & Llama, Qwen \\
mole & The spy was a mole in the agency & She had a small mole on her cheek & Qwen \\
nail & Hammer the nail into the wood & She painted each nail bright red & Qwen \\
novel & She wrote a fascinating novel & His idea was a novel solution & Llama, Qwen \\
organ & The heart is a very important organ & She learned to play the church organ & Llama, Qwen \\
pen & The blue pen writes very smoothly & The large pen holds many pigs & Llama, Qwen \\
pick & She used a pick to open the lock & You can now pick your favorite color & GPT-2, Llama, Qwen \\
pitcher & The glass pitcher holds cold water & The star pitcher throws fast balls & Qwen \\
plane & The passenger plane took off early & Use a plane to smooth wood & GPT-2, Llama, Qwen \\
plant & The green plant needs more water & The factory plant operates all day & Qwen \\
plate & Put the food on a plate & The car has a California plate &  \\
play & The children love to play here & They watched a great play & Qwen \\
plot & They uncovered a secret plot & She bought a small plot of land & Qwen \\
point & You have a very good point & The pencil has a sharp point &  \\
produce & That factory can produce many cars & We buy fresh produce at the market & GPT-2, Llama, Qwen \\
range & This cooker has a wide range & The mountains form a high range &  \\
rank & He holds a very high rank & The old meat smells quite rank &  \\
rare & That old coin is very rare & She likes her steak cooked rare &  \\
resume & Submit your resume with the form & We will resume the meeting later & GPT-2, Llama, Qwen \\
right & You have the right to remain silent & Please make a right turn up ahead & Llama, Qwen \\
ring & The gold ring fits her finger & The large ring contains the boxers &  \\
rock & The heavy rock sits on ground & The loud rock plays through speakers &  \\
rose & The beautiful red rose smells lovely & The morning sun rose very quickly & GPT-2, Llama, Qwen \\
second & Wait just one second before leaving & She finished in second place today & GPT-2, Llama, Qwen \\
sentence & Write a complete sentence with care & The judge will sentence defendant soon & GPT-2, Llama, Qwen \\
sewer & The tailor is an expert sewer & The city fixed the main sewer & GPT-2, Llama, Qwen \\
shop & The bakery shop opens at six & She loves to shop for clothes &  \\
skip & You should not skip your breakfast & The happy children skip down the path & Llama, Qwen \\
slip & Be careful not to slip here & She wrote on a slip paper & GPT-2, Llama \\
sow & The large mother sow is protective & The farmer will sow the seeds soon & GPT-2, Llama, Qwen \\
spot & This is a perfect spot for a picnic & She saw a tiny spot of ink on the page & GPT-2, Qwen \\
spring & The warm spring brings new flowers & The cold spring provides fresh water & Qwen \\
stamp & Put a stamp on the envelope & She will stamp her foot angrily & GPT-2, Llama, Qwen \\
stick & The wet mud will stick to your shoes & He threw the long stick for his dog & GPT-2, Qwen \\
strike & The workers will strike for wages & The clock will strike twelve soon & GPT-2, Llama, Qwen \\
subject & The king will subject them to his rule & Her favorite school subject is history & GPT-2, Llama, Qwen \\
suit & He wore a dark business suit & The lawyer filed a civil suit & GPT-2, Llama, Qwen \\
tear & A single sad tear ran down her face & You might easily tear the thin paper & GPT-2, Llama, Qwen \\
tender & The company submitted a tender & This steak is extremely tender & GPT-2, Llama, Qwen \\
tick & He found a small tick on the dog & She put a red tick by the answer &  \\
toast & Let's raise a toast to their success & She likes to eat toast for breakfast &  \\
train & The passenger train departs at three & She will train for the marathon & GPT-2, Llama, Qwen \\
trip & Be careful not to trip on the rug & They are planning a trip to the sea & GPT-2, Llama, Qwen \\
trunk & The car trunk holds heavy luggage & The tree trunk grows very thick &  \\
watch & She loves her new watch & They will sit and watch & Llama, Qwen \\
well & The deep well provides clean water & She sings well at every concert & GPT-2, Llama, Qwen \\
will & She will see you tomorrow morning & Her will is very strong indeed & Llama, Qwen \\
wind & You need to wind the old clock & The strong autumn wind is blowing & GPT-2, Llama, Qwen \\
yard & The children played in the yard & The fabric is sold by yard & GPT-2 \\
yield & The farm will yield good harvest & Drivers must yield to traffic here & GPT-2 \\
\end{longtable}
}
\clearpage
\noindent{\bfseries Table S4}\par
\noindent Activation patching concept-drift polyseme stimuli: position-matched candidate pool of concept-drift pairs, prior to per-model validation\par\smallskip
{\fontsize{6.7}{8.0}\selectfont
\begin{longtable}{@{}>{\raggedright\arraybackslash}p{0.10\linewidth}>{\raggedright\arraybackslash}p{0.32\linewidth}>{\raggedright\arraybackslash}p{0.32\linewidth}>{\raggedright\arraybackslash}p{0.18\linewidth}@{}}
\toprule
\textbf{Word} & \textbf{Context 1} & \textbf{Context 2} & \textbf{Models used} \\
\midrule
\endfirsthead
\multicolumn{4}{l}{\textit{Table S4 continued}} \\
\toprule
\textbf{Word} & \textbf{Context 1} & \textbf{Context 2} & \textbf{Models used} \\
\midrule
\endhead
\midrule
\multicolumn{4}{r}{\textit{Continued on next page}} \\
\endfoot
\bottomrule
\endlastfoot
address & Write your home address on the form & Type the web address in the browser & GPT-2, Llama, Qwen \\
boot & She wore a leather boot on her foot & System needs a quick boot for updates & Llama \\
bounce & The tennis ball can bounce over the high net & We should probably now bounce from this boring party &  \\
bug & The small bug crawls on the leaf & The software bug crashes the program &  \\
cancel & We had to cancel our dinner reservation & They tried to cancel the celebrity online &  \\
cap & He wore a baseball cap to the game & That story is complete cap if honest & GPT-2, Llama, Qwen \\
chip & He found a wood chip on the floor & Computer needs a silicon chip for power & Qwen \\
cloud & The dark cloud brings heavy rain & The data cloud stores our photos & Llama \\
cookie & She likes a warm cookie with her milk & This website needs a cookie for login & Qwen \\
crash & The terrible car crash blocked the road & The sudden system crash deleted the files & Qwen \\
crush & She tried not to crush the flowers & She admitted to her crush on him & GPT-2, Llama, Qwen \\
drive & He has a strong drive to always succeed & He has a flash drive to store his files & Qwen \\
epic & They read an ancient Greek epic in class & That concert was a truly epic experience & Llama, Qwen \\
feed & They need to gently feed the baby birds & They check their news feed for the updates & GPT-2, Llama, Qwen \\
file & Use the metal file to smooth edges & Save the computer file to the desktop &  \\
filter & Pour coffee through the paper filter now & She applied a beautiful photo filter today &  \\
fire & The burning fire warmed the cold room & That new fire song played all night & GPT-2, Llama, Qwen \\
flame & The candle had a steady flame & His post was a deliberate flame & GPT-2, Llama, Qwen \\
follow & Please follow me to the conference room & Please follow my account on Twitter &  \\
footprint & We saw a large bear footprint today & You leave a unique digital footprint online & Llama, Qwen \\
fresh & We bought some very fresh produce today & Those shoes look really fresh to me &  \\
friend & She is a very good friend to many people & You should quickly go and friend those people & GPT-2, Llama, Qwen \\
ghost & The scary ghost haunts the mansion & He will ghost her after dates & GPT-2, Llama, Qwen \\
hack & They used an axe to hack the wood & They tried their best to hack the server & Llama \\
handle & Turn the door handle to open it & Choose your Twitter handle very carefully & GPT-2 \\
host & The gracious host welcomed all guests & The web host handles all requests & Llama, Qwen \\
keyboard & The piano keyboard has eighty-eight keys & My computer keyboard needs new batteries &  \\
link & The broken chain link fell to ground & The website page link leads to home & GPT-2, Llama, Qwen \\
lit & She has now safely lit the candle & That party was very lit last night & GPT-2, Llama, Qwen \\
mad & He was so incredibly mad at his friend & He has some really mad skills on the board & GPT-2, Llama, Qwen \\
memory & Her childhood memory was very vivid & The computer memory is running low &  \\
menu & Read the restaurant menu very carefully & Open the context menu with right click & Qwen \\
mouse & The small mouse runs very quickly & The wireless mouse works very smoothly &  \\
page & Turn to book page fifty very quickly & Load the web page slowly for me & Llama, Qwen \\
paste & Apply the adhesive paste to the wall & Just copy and paste the copied text & GPT-2, Llama, Qwen \\
platform & The train platform was very crowded & The social platform is extremely popular &  \\
portal & The stone portal led into the castle & The student portal allows easy access &  \\
post & The wooden fence post stands quite tall & Her social media post received many likes &  \\
profile & She updated her user profile this morning & The artist drew her profile from the side &  \\
screen & The privacy screen blocks prying eyes & My computer screen is too bright & GPT-2, Llama, Qwen \\
server & The restaurant server brought our food & The web server processes our data & GPT-2, Llama \\
shade & They sat in the welcome shade of the tree & She threw some serious online shade at him & Llama, Qwen \\
share & Children should share toys with others & Users can share posts with friends & GPT-2 \\
ship & The cargo ship sailed across the ocean & I totally ship those two characters & GPT-2, Llama, Qwen \\
sick & The poor child felt very sick today & Those new tricks looked very sick indeed &  \\
site & The construction site is very busy & The news site provides daily updates & GPT-2, Llama \\
slick & That was a really slick move last night & The wet road was slick from the rain & Llama, Qwen \\
smart & She is very smart at mathematics & This is my smart TV for streaming & GPT-2, Llama, Qwen \\
snack & I think I will have a small snack now & My friends think he is such a snack now & GPT-2, Llama \\
spam & We ate fried spam for our breakfast & I delete unwanted spam from my inbox & Llama, Qwen \\
status & Her social status improved significantly & Her online status shows she's available &  \\
stream & The mountain stream flows through valleys & The video stream plays through internet & Qwen \\
surf & I love to surf big ocean waves & I like to surf the internet daily &  \\
tablet & Take the tablet with clean water & Use the tablet for reading books & Qwen \\
tag & He removed the price tag from the jacket & He was careful to tag his friends online & GPT-2, Llama, Qwen \\
terminal & The bus terminal was very crowded & The computer terminal runs all commands & Qwen \\
thread & The sewing thread is very strong & The forum thread is quite interesting & Qwen \\
timeline & The history timeline shows major events & Her social timeline displays recent posts & Llama, Qwen \\
trash & Please take out the garbage trash tonight & They are going to completely trash the place & GPT-2, Llama, Qwen \\
tweet & The bird tweet sounds so sweet & The Twitter tweet received many likes &  \\
vibe & The strange sound created an odd vibe & This coffee shop has a nice vibe &  \\
viral & The viral infection spreads very rapidly & The viral video spreads on Twitter &  \\
virus & The dangerous virus caused severe illness & The computer virus infected the network &  \\
wall & They painted the bedroom wall last night & She updated the Facebook wall last night & Qwen \\
wild & They saw many strange and wild animals there & That concert was absolutely and wild last night & GPT-2, Llama, Qwen \\
shook & The sudden tremor shook the building & The plot twist really shook the audience & GPT-2 \\
code & The spy used a secret code carefully & She writes the Python computer code every day &  \\
bomb & They found a time bomb in the basement & That new restaurant is the bomb you should try it &  \\
window & Open the bedroom window for fresh air & Close the browser window when you finish & GPT-2 \\
\end{longtable}
}
\clearpage
\subsection*{Sequence-order control stimuli}
Table S5 lists the sentence pairs used in the Llama-3.2-3B sequence-order control analysis. Each pair contains overlapping lexical material arranged in a different order or semantic-role configuration, allowing contextual and positional variation to be compared with the separation observed for ambiguous-word pairs.
\smallskip
\\
\noindent{\bfseries Table S5}\par
\noindent Sequence-order control sentence pairs used in the Llama-3.2-3B control analysis\par\smallskip
{\fontsize{7.0}{8.3}\selectfont
\begin{longtable}{@{}>{\raggedright\arraybackslash}p{0.18\linewidth}>{\raggedright\arraybackslash}p{0.37\linewidth}>{\raggedright\arraybackslash}p{0.37\linewidth}@{}}
\toprule
\textbf{Control type} & \textbf{Sentence 1} & \textbf{Sentence 2} \\
\midrule
\endfirsthead
\multicolumn{3}{l}{\textit{Table S5 continued}} \\
\toprule
\textbf{Control type} & \textbf{Sentence 1} & \textbf{Sentence 2} \\
\midrule
\endhead
\midrule
\multicolumn{3}{r}{\textit{Continued on next page}} \\
\endfoot
\bottomrule
\endlastfoot
Agent-Patient Reversals & The cat chased the mouse. & The mouse chased the cat. \\
 & The teacher taught the student. & The student taught the teacher. \\
 & The dog bit the mailman. & The mailman bit the dog. \\
 & The lion hunted the zebra. & The zebra hunted the lion. \\
 & The doctor examined the patient. & The patient examined the doctor. \\
 & The lawyer questioned the witness. & The witness questioned the lawyer. \\
 & The chef trained the apprentice. & The apprentice trained the chef. \\
 & The writer inspired the reader. & The reader inspired the writer. \\
 & The parent guided the child. & The child guided the parent. \\
 & The director filmed the actor. & The actor filmed the director. \\
Source-Destination / Container Reversals & I put the dishes in the cabinet. & I put the cabinet in the dishes. \\
 & She poured the water into the glass. & She poured the glass into the water. \\
 & He placed the book on the shelf. & He placed the shelf on the book. \\
 & They loaded the boxes onto the truck. & They loaded the truck onto the boxes. \\
 & I stuck the stamp on the envelope. & I stuck the envelope on the stamp. \\
 & She hung the picture on the wall. & She hung the wall on the picture. \\
 & He put the keys in the drawer. & He put the drawer in the keys. \\
 & They stored the files in the cabinet. & They stored the cabinet in the files. \\
 & I placed the flowers in the vase. & I placed the vase in the flowers. \\
 & She laid the blanket on the bed. & She laid the bed on the blanket. \\
Ingredient-Container / Addition Order Reversals & Add the eggs to the flour. & Add the flour to the eggs. \\
 & The chef added salt to the soup. & The chef added soup to the salt. \\
 & Mix the sugar into the batter. & Mix the batter into the sugar. \\
 & Pour the milk into the coffee. & Pour the coffee into the milk. \\
 & Stir the cream into the sauce. & Stir the sauce into the cream. \\
 & Add the spices to the mixture. & Add the mixture to the spices. \\
 & Blend the fruit into the smoothie. & Blend the smoothie into the fruit. \\
 & Fold the chocolate into the dough. & Fold the dough into the chocolate. \\
Possession / Transfer Reversals & She gave a book to him. & She gave him a book. \\
 & John lent Mary the car. & Mary lent John the car. \\
 & The company sent the client a package. & The client sent the company a package. \\
 & I handed the waiter the menu. & The waiter handed me the menu. \\
 & She passed him the salt. & He passed her the salt. \\
 & The bank loaned the customer money. & The customer loaned the bank money. \\
 & I showed my friend the photo. & My friend showed me the photo. \\
 & The artist sold the collector a painting. & The collector sold the artist a painting. \\
Temporal/Sequential Reversals & She ate breakfast before work. & She ate work before breakfast. \\
 & He finished the homework after dinner. & He finished dinner after homework. \\
 & They arrived before the show started. & The show started before they arrived. \\
 & I cleaned the house after the party. & I cleaned the party after the house. \\
Modifier Scope / Attachment Differences & The old man's hat was red. & The man's old hat was red. \\
 & She saw the boy with binoculars. & She saw the boy with the binoculars. \\
 & The tall girl's brother arrived. & The girl's tall brother arrived. \\
 & He bought the expensive car's insurance. & He bought the car's expensive insurance. \\
Semantic Role Reversals (various) & The key opened the door. & The door opened the key. \\
 & The hammer broke the window. & The window broke the hammer. \\
 & Fire melted the ice. & Ice melted the fire. \\
 & The sun dried the clothes. & The clothes dried the sun. \\
 & Wind scattered the leaves. & Leaves scattered the wind. \\
 & The knife cut the bread. & The bread cut the knife. \\
 & Rain filled the bucket. & The bucket filled the rain. \\
 & The magnet attracted the metal. & The metal attracted the magnet. \\
Idiomatic / Semantic Anomalies & Time flies like an arrow. & Flies like time like an arrow. \\
 & The grass is always greener. & Always greener is the grass. \\
 & Birds of a feather flock together. & Feather of birds flock together. \\
 & Actions speak louder than words. & Words speak louder than actions. \\
\end{longtable}
}
\clearpage
\subsection*{Protected foundation-size analysis}
To assess whether protecting a larger initial foundation altered the single-layer disruption results, analyses were repeated with foundation sizes of 2, 3, and 4. A foundation size of $k$ indicates that the first $k$ layers were protected and that disruption was applied only at subsequent eligible layers. Foundation-size effects were evaluated separately for each model, ambiguity type, and degradation metric using Kruskal--Wallis tests. These secondary analyses used the raw degradation measures underlying the percentage-degradation visualisations in the main text.
\smallskip
\\
\noindent{\bfseries Table S6}\par
\noindent Omnibus tests of foundation-size effects on degradation\par\smallskip
{\fontsize{7.2}{8.5}\selectfont
\begin{longtable}{@{}>{\raggedright\arraybackslash}p{0.11\linewidth}>{\raggedright\arraybackslash}p{0.15\linewidth}>{\raggedright\arraybackslash}p{0.30\linewidth}rrr@{}}
\toprule
\textbf{Model} & \textbf{Ambiguity type} & \textbf{Metric} & \textbf{$H$} & \textbf{df} & \textbf{$p$} \\
\midrule
\endfirsthead
\multicolumn{6}{l}{\textit{Table S6 continued}} \\
\toprule
\textbf{Model} & \textbf{Ambiguity type} & \textbf{Metric} & \textbf{$H$} & \textbf{df} & \textbf{$p$} \\
\midrule
\endhead
\midrule
\multicolumn{6}{r}{\textit{Continued on next page}} \\
\endfoot
\bottomrule
\endlastfoot
GPT-2 & Homonyms & Activation distance & 0.49 & 2 & 0.7838 \\
GPT-2 & Homonyms & Logit distance & 0.22 & 2 & 0.8978 \\
GPT-2 & Homonyms & KL divergence & 7.86 & 2 & \textbf{0.0196} \\
GPT-2 & Concept drift & Activation distance & 1.91 & 2 & 0.3855 \\
GPT-2 & Concept drift & Logit distance & 0.43 & 2 & 0.8073 \\
GPT-2 & Concept drift & KL divergence & 3.98 & 2 & 0.1369 \\
Llama & Homonyms & Activation distance & 19.65 & 2 & \textbf{5.42e-5} \\
Llama & Homonyms & Logit distance & 43.51 & 2 & \textbf{3.56e-10} \\
Llama & Homonyms & KL divergence & 81.55 & 2 & \textbf{1.96e-18} \\
Llama & Concept drift & Activation distance & 4.80 & 2 & 0.0909 \\
Llama & Concept drift & Logit distance & 4.00 & 2 & 0.1351 \\
Llama & Concept drift & KL divergence & 21.03 & 2 & \textbf{2.71e-5} \\
Qwen & Homonyms & Activation distance & 0.45 & 2 & 0.7982 \\
Qwen & Homonyms & Logit distance & 20.82 & 2 & \textbf{3.01e-5} \\
Qwen & Homonyms & KL divergence & 5.23 & 2 & 0.0733 \\
Qwen & Concept drift & Activation distance & 0.76 & 2 & 0.6833 \\
Qwen & Concept drift & Logit distance & 14.16 & 2 & \textbf{8.42e-4} \\
Qwen & Concept drift & KL divergence & 1.15 & 2 & 0.5633 \\
\end{longtable}
}
\noindent\textit{Note.} Bold $p$ values indicate $p<.05$ without correction across the 18 omnibus tests. Foundation-size analyses were exploratory secondary analyses.
\medskip
\\
\noindent{\bfseries Table S7}\par
\noindent Foundation-size descriptives and significant Holm-adjusted post-hoc contrasts for metrics with significant omnibus tests\par\smallskip
{\fontsize{6.8}{8.0}\selectfont
\begin{longtable}{@{}>{\raggedright\arraybackslash}p{0.08\linewidth}>{\raggedright\arraybackslash}p{0.12\linewidth}>{\raggedright\arraybackslash}p{0.17\linewidth}rrr>{\raggedright\arraybackslash}p{0.31\linewidth}@{}}
\toprule
\textbf{Model} & \textbf{Type} & \textbf{Metric} & \textbf{F2} & \textbf{F3} & \textbf{F4} & \textbf{Significant contrasts} \\
\midrule
\endfirsthead
\multicolumn{7}{l}{\textit{Table S7 continued}} \\
\toprule
\textbf{Model} & \textbf{Type} & \textbf{Metric} & \textbf{F2} & \textbf{F3} & \textbf{F4} & \textbf{Significant contrasts} \\
\midrule
\endhead
\midrule
\multicolumn{7}{r}{\textit{Continued on next page}} \\
\endfoot
\bottomrule
\endlastfoot
GPT-2 & Homonyms & KL divergence & 12.4364 & 12.8072 & 13.1696 & F2 < F4 ($r_{rb}=-0.093$, $p_H=0.0190$) \\
Llama & Homonyms & Activation distance & 0.0378 & 0.0405 & 0.0446 & F2 < F4 ($r_{rb}=-0.066$, $p_H=3.31e-5$); F3 < F4 ($r_{rb}=-0.040$, $p_H=0.0168$) \\
Llama & Homonyms & Logit distance & 0.0590 & 0.0603 & 0.0656 & F2 < F4 ($r_{rb}=-0.093$, $p_H=2.10e-9$); F3 < F4 ($r_{rb}=-0.079$, $p_H=5.00e-7$) \\
Llama & Homonyms & KL divergence & 0.6786 & 0.7035 & 0.7294 & F2 < F3 ($r_{rb}=-0.068$, $p_H=9.17e-6$); F2 < F4 ($r_{rb}=-0.136$, $p_H=7.89e-19$); F3 < F4 ($r_{rb}=-0.070$, $p_H=9.17e-6$) \\
Llama & Concept drift & KL divergence & 0.7192 & 0.7477 & 0.7712 & F2 < F3 ($r_{rb}=-0.050$, $p_H=0.0314$); F2 < F4 ($r_{rb}=-0.093$, $p_H=2.41e-5$); F3 < F4 ($r_{rb}=-0.051$, $p_H=0.0314$) \\
Qwen & Homonyms & Logit distance & 0.1756 & 0.1864 & 0.1354 & F2 < F3 ($r_{rb}=-0.042$, $p_H=9.70e-4$); F3 > F4 ($r_{rb}=0.051$, $p_H=8.84e-5$) \\
Qwen & Concept drift & Logit distance & 0.1801 & 0.1932 & 0.1265 & F2 > F4 ($r_{rb}=0.045$, $p_H=0.0103$); F3 > F4 ($r_{rb}=0.057$, $p_H=0.0014$) \\
\end{longtable}
}
\noindent\textit{Note.} F2--F4 are median raw degradation values for foundation sizes 2--4. Only pairwise contrasts surviving Holm correction within each model-by-domain-by-metric family are listed. $r_{rb}$ is the signed rank-biserial correlation for the first foundation relative to the second; negative values indicate greater degradation for the second foundation.
\medskip
Foundation size had limited and model-dependent effects. In Llama, larger protected foundations were associated with greater degradation for all three homonym metrics and for concept-drift KL divergence; KL degradation increased monotonically across foundation sizes 2, 3, and 4. GPT-2 showed only a small increase in homonym KL degradation between foundations 2 and 4. Qwen showed non-monotonic effects: homonym logit-distance degradation peaked at foundation 3, whereas concept-drift logit-distance degradation was lower at foundation 4 than at foundations 2 and 3. Thus, the protected foundation influenced measured degradation in selected analyses, but the direction and magnitude of the effect were architecture- and metric-specific.
\bibliographystyle{compling}